\documentclass[11pt]{article}
\usepackage{comment}
\usepackage[utf8]{inputenc}
\usepackage[T1]{fontenc}
\usepackage{lmodern}
\usepackage[margin=1in]{geometry}
\usepackage{graphicx}
\usepackage{subcaption}
\usepackage{amsmath}
\usepackage{microtype}
\usepackage{setspace}
\usepackage[super,comma,sort&compress]{natbib}
\usepackage[hidelinks]{hyperref}
\usepackage{caption}
\usepackage{newfloat} \DeclareFloatingEnvironment[ fileext=loe, listname={Extended Data Figures}, name={Extended Data Figure}, placement=htbp ]{extendedfigure} 

\title{\bfseries High-Stakes Decisions with Language Models:\\Insights from Emergency Triage}

\author{Khurram Yamin$^{1,2}$\thanks{Research performed during an internship at Microsoft.}  \ 
\ Christopher Kelly$^{1}$  \  \ Bryan Wilder$^{2}$ \ \ Eric Horvitz$^{1}$\thanks{Correspondence: \texttt{horvitz@microsoft.com}.} \\[0.6em] {\normalsize $^{1}$Microsoft } \\ $^{2}$Carnegie Mellon University }

\date{July 28, 2026}

\begin{document}

\maketitle

\begin{abstract}
\noindent
High-stakes decisions under uncertainty, such as medical emergency triage, require more than accurate predictions. They depend on estimating the likelihood of alternative outcomes while explicitly weighing the consequences of different actions, principles that have long formed the foundation of medical diagnosis and decision making. Yet language models are increasingly used for high-stakes clinical recommendations without explicit specification of the utilities governing these decisions. Here we show that emergency triage with language models can be understood within a probabilistic decision framework, providing a case study of a broader decision-analytic paradigm for steering, evaluating, and deploying language models in high-stakes settings. Using clinical vignettes from a structured evaluation of a consumer triage system, we analyze recommendations for treatment under alternative utility functions that specify the relative costs of missed emergencies and unnecessary escalation. We find that capable language models adjust recommendations in response to stated utilities, revealing that the same underlying predictions can support markedly different decision policies. These findings show that effective deployment depends not only on improving predictions but also on making decision objectives explicit. More broadly, they suggest that language models for high-stakes applications should be understood and evaluated as probabilistic decision systems whose recommendations depend jointly on predictive performance and explicit utilities.
\end{abstract}

\section*{Introduction}

Although the performance of frontier language models (LMs) on medical reasoning tasks has improved, studies continue to document brittleness and dangerous failures.\cite{Gu2026,navarro2026internalrepresentationclinicalknowledge,Kim2025,Gaber2025,Williams2024edllm,Auger2026.03.23.26349082,Masanneck2024TriagePerformance} A salient example comes from a recent investigation by Ramaswamy et al.~(2026).\cite{ramaswamy2026chatgpthealth} They showed that ChatGPT Health, a consumer health system built on a frontier LM and released in early 2026, failed to appropriately recommend emergency-department evaluation for patients with life-threatening conditions. The authors found that the system under-triaged more than half of the gold-standard emergencies. Patients with conditions such as diabetic ketoacidosis or impending respiratory failure were advised to seek evaluation within one to two days rather than to seek immediate emergency care.  Understanding why such failures occur is essential for designing trustworthy decision-support systems.  Emergency medical triage is a canonical high-stakes decision problem: recommendations must be made under uncertainty, errors have asymmetric consequences, and the appropriate trade-off depends on the decision-maker's priorities. We use medical triage to address a broader question: how should language models be designed and evaluated when translating uncertainty into consequential decisions? 

A natural interpretation is that failures, such as those observed by Ramaswamy et al.~(2026), reflect inadequate diagnostic competence. We revisit the dataset, methods, and results to argue that emergency triage is not simply a prediction problem but a decision problem. Understanding this distinction requires separating the two components of a decision under uncertainty. The first is a probabilistic assessment of the state of the world: what is likely to be true? The second is a utility judgment specifying how the consequences of different actions should be valued. Neither component alone is sufficient for making good decisions. In triage, the relevant probability is that the patient requires emergency care. The relevant utility judgment is how strongly to prioritize avoiding missed emergencies over unnecessary emergency-department referrals. The distinction between probabilities and utilities is central to decision theory\cite{VonNeumann1944-VONTOG-4,savage1972foundations,horvitz1988decision} and  clinical decision analysis.\cite{10.1371/journal.pone.0109264} Two clinicians may agree on the probability of an emergency yet recommend different actions because they assign different relative costs to false negatives and false positives. Clinical protocols already encode such choices: stroke, sepsis, and acute cardiac-event policies deliberately tolerate false alarms to avoid missed emergencies. LM triage systems must make the same trade-off, but they generally do so without explicitly specifying how missed emergencies should be weighed against unnecessary emergency-referrals. 

In this framing, triage failures may reflect poor clinical inference, inappropriate weighting of these harms, or both. The appropriate remedy depends on the source of the failure. If a model cannot recognize an emergency, it needs better diagnostic capabilities. If it applies inappropriate utilities, it must instead be guided by explicit priorities and steered toward the desired trade-off. Evaluations that measure only a model's default behavior cannot distinguish these cases.  Importantly, there is no universally correct utility function. A crowded urban emergency department, a remote clinic, an anxious parent at midnight, and a public-health payer may reasonably prefer different trade-offs. When a model is asked to triage without explicitly specified utilities, its actions necessarily implement an implicit trade-off between the costs of different errors. Whether these implicit utilities are appropriate cannot be assessed and steered unless they are made explicit. There is little reason to expect them to be aligned with the preferences of clinicians, patients, or health care organizations. \cite{mazeika2025utilityengineeringanalyzingcontrolling,slama2026llmpreferencespredictdownstream,liu2026generativevalueconflictsreveal,zhu2025steeringriskpreferenceslarge,ouyang2025aidecisionmakerethicsrisk}

\begin{figure}[htbp]
  \centering
  \includegraphics[width=.9\linewidth]{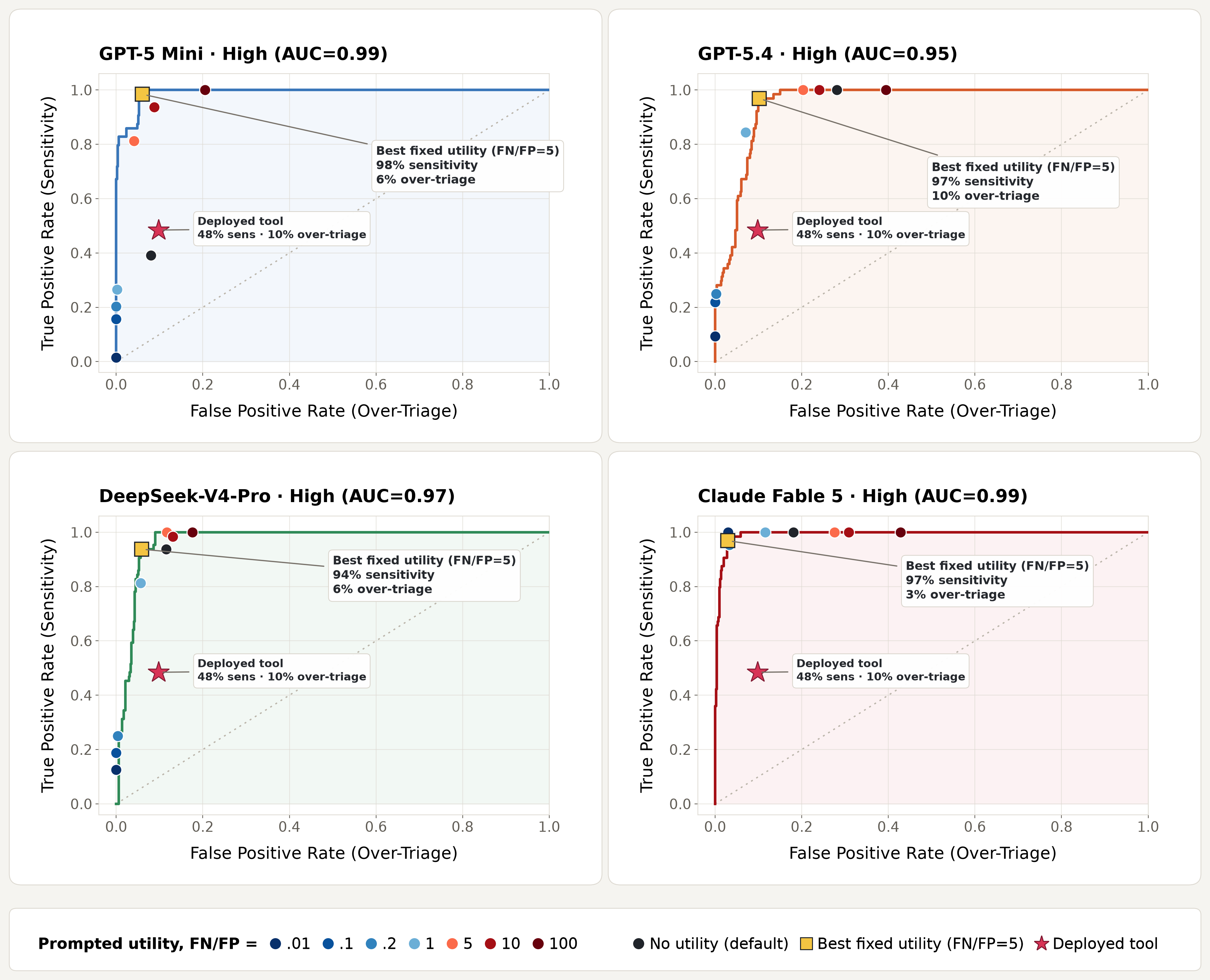}
  \caption{\textbf{Emergency triage performance and safety-resource trade-offs.} Curves trace the safety--resource trade-off implied by the elicited
  probabilities of GPT-5-mini (reported as the backbone of ChatGPT Health),
  GPT-5.4, DeepSeek-V4-Pro, and Claude Fable 5 (each evaluated with high reasoning). The red star marks the 
  behavior of ChatGPT Health documented by Ramaswamy et al.~(2026), which identified only 48\% of emergencies
  while erroneously referring approximately 10\% of non-emergency cases, indicating a strong prioritization of resource conservation over safety. Black circles show each model’s operating point when no utility trade-off is specified. Colored circles
  show the operating points induced by explicit cost-ratio prompts, demonstrating that utility-based prompting can steer models across a range of safety-resource tradeoffs. The yellow square marks the optimal fixed-threshold operating point under a
  safety-weighted 5:1 cost ratio---that is the probability threshold that minimizes
  cost-weighted error against the gold-standard clinical labels when a missed
  emergency is penalized five times as heavily as an unnecessary referral. Other
  cost ratios select different points along the same curve, as shown in
  Extended Data Fig.~\ref{extfig:roc} where we show 18 model variants.}
  \label{fig:original}
\end{figure}

\begin{figure}[htbp]
  \centering
  \includegraphics[width=.9\linewidth]{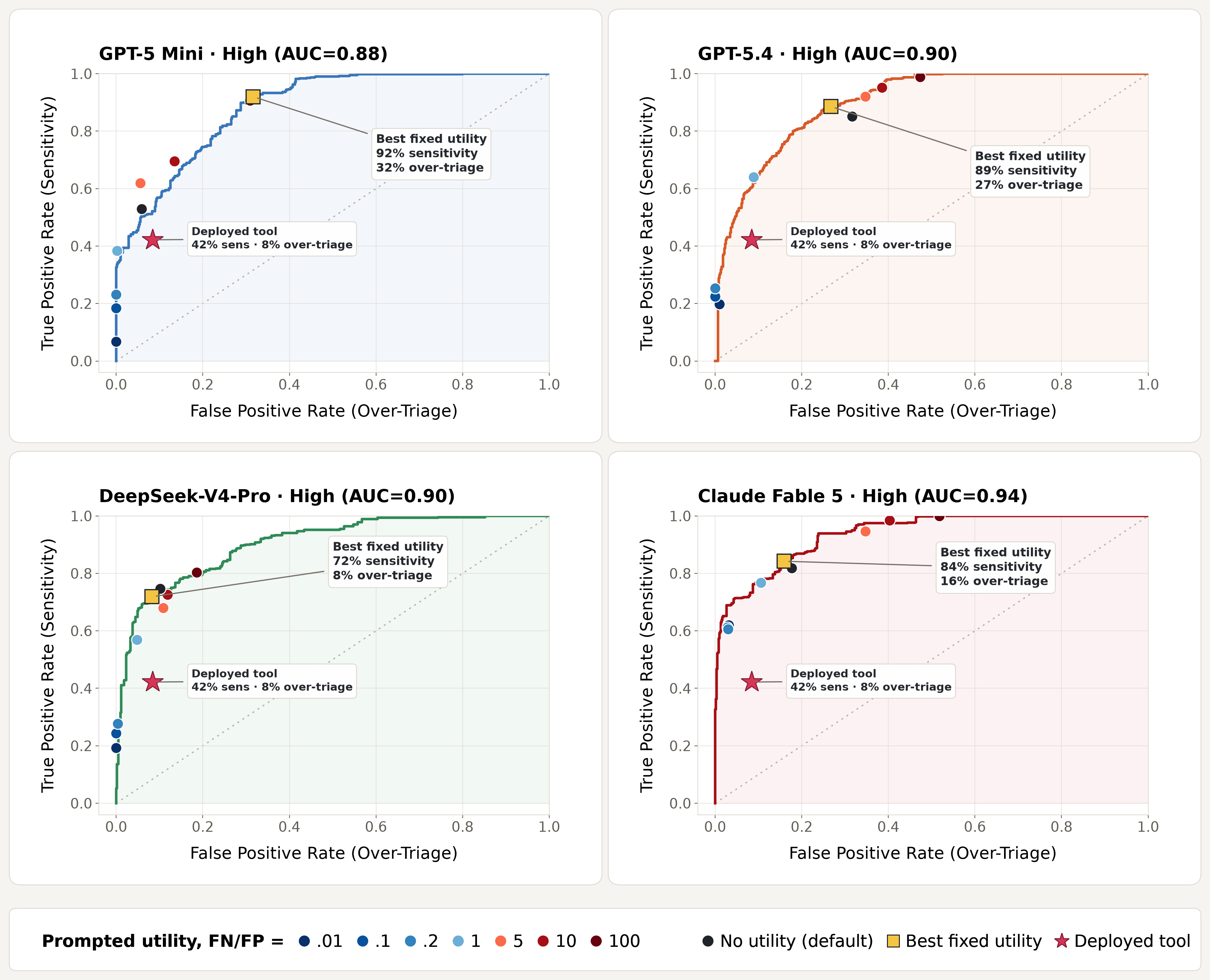}
  \caption{\textbf{Emergency triage performance for the expanded case set.} Triage performance when cases classified as ``edge'' cases by Ramaswamy et al.~(2026) are included in the emergency category. Plot elements and the optimal fixed-threshold operating point under the 5:1 cost ratio are defined as in Figure~\ref{fig:original}. Using this expanded emergency-care classification, GPT-5-mini  identifies 42\% of emergencies while referring 8\% of non-emergency cases, indicating a strong prioritization of resource conservation over safety. Across the models evaluated with high-reasoning, AUROCs range from 0.88 to 0.94.}
  \label{fig:expanded}
\end{figure}

In this paper, we distinguish errors arising from probabilistic assessment from those arising from utility judgments in the clinical vignettes studied by Ramaswamy et al.~(2026), building on recent work that shows how the implicit utilities underlying language model choices can be inferred.\cite{yamin2026agentssaythinganother,yamin2026revealedpreferencesclarifyllm} We further show that a model's implicit triage utilities can be recovered from its behavior, that much of the deployed tool's unsafe default behavior can be explained by the absence of an explicit specification of the relative priority assigned to avoiding under-triage versus over-triage, and that plain-language utility instructions shift the behavior of capable models toward specified clinical priorities. Together, these analyses provide a framework for distinguishing limitations in clinical inference from failures arising from miscalibration or the application of inappropriate implicit utilities. They instantiate a broader decision-analytic paradigm for model steering, evaluation, and deployment, in which probabilistic assessment, calibration, utilities, and the mapping from predictions to actions are made explicit. We refer to methods that guide model behavior toward specified utilities or operating points as \emph{decision-analytic steering}, and to the broader goal of aligning recommendations with explicit decision objectives as \emph{decision-analytic alignment}. Although we focus on emergency triage, these methods apply to a broad range of decision contexts in which errors have asymmetric consequences and the appropriate trade-off depends on the decision-maker's priorities. We conclude with recommendations for prompting, evaluating, and designing LM systems for triage and other high-stakes decision settings. 

\section*{Decoding a model's decision priorities}

A triage decision about whether to recommend emergency care can
go wrong in two ways: a \emph{false negative} occurs when a patient who needs emergency care is under-triaged, whereas a \emph{false positive} occurs when a patient who does not need emergency care is over-triaged. The 
central value judgment in triage is how to trade off these two errors. A model's handling of this trade-off can be summarized by a \emph{cost ratio}: the cost assigned to a false negative (a
missed emergency) relative to that assigned to a false positive (an unnecessary referral). A
cost ratio of ten means that the model behaves as though missing a true emergency is
ten times as costly as an unnecessary visit, reflecting a strongly safety-prioritized stance. A cost
ratio well below one indicates the opposite: the model behaves as though avoiding unnecessary visits matters more than catching every emergency, reflecting a resource-prioritized stance. Each cost ratio defines a probability threshold for referral---the estimated likelihood that emergency care is needed above which the model should recommend emergency care. If missed emergencies are much more costly than unnecessary referrals, the model should recommend emergency care even when the estimated probability that emergency care is needed is relatively low. If unnecessary referrals are weighted more heavily, the model should require a higher estimated probability before recommending emergency care.   

We analyze a set of medical vignettes by eliciting probability estimates separately from triage decisions.  For each vignette, we first ask the model to estimate the probability that the patient needs emergency care. We treat this estimate as the model's expressed probability.  In a separate query, we ask the model to provide a triage recommendation. We treat this response as its decision. Jointly analyzing a language model's probability estimates and decisions enables us to recover the cost ratio that best explains its behavior across cases. A model that refers patients even when its own estimated probability that emergency care is needed is low behaves as though it places a high relative cost on missed emergencies. A model that withholds referral until its estimated probability is high behaves as though it places greater relative weight on avoiding unnecessary referrals.

We then test whether these priorities can be controlled through prompting. We instruct the model to adopt specified cost ratios and measure whether its decisions shift accordingly. This black-box procedure requires no access to model internals; it uses only probability estimates and triage recommendations elicited through prompting. Full details are provided in Methods. We apply this framework to the clinical vignettes from the triage stress test described above, using multiple  frontier models evaluated at several reasoning settings.


We first revisit a key finding reported by Ramaswamy et al.~(2026): the poor triage performance of the ChatGPT Health consumer-facing service. At the time of the study, the service was documented as using GPT-5-mini with reasoning.\cite{ramaswamy2026chatgpthealth} We therefore compared its published triage decisions with the trade-off curve implied by GPT-5-mini's estimated probabilities. 

Following Ramaswamy et al. (2026), our primary analysis included 576 of the original 960 cases, which were generated through systematic variations of physician-authored vignettes. The  study had excluded 384 ``edge'' cases in which physicians judged that the patient required urgent medical attention but were uncertain whether immediate emergency-department care was necessary. Of the 576 included cases, 64 were classified as emergencies because adjudicating physicians unambiguously concluded that immediate emergency-department care was required; the remaining 512 were classified as non-emergencies. The results in Figure~\ref{fig:original} shows emergency-triage performance and the resulting safety--resource trade-offs across the models evaluated.

To assess the generality of the methodology, we conducted a second analysis using the expanded set of 1,248 cases from Ramaswamy et al.~(2026). The expanded set comprised the 960 original case variants and an additional 288 textbook cases. Under an expanded definition of emergency, 640 case variants were classified as requiring emergency care. These included 192 cases for which physicians unambiguously judged immediate emergency-department care to be necessary (64 original emergency cases and 128 textbook cases) and 448 cases in which physicians judged that urgent attention was necessary but were uncertain whether the patient should be sent to the emergency department or evaluated by a physician within 24--48 hours (384 original edge cases and 64 textbook variants). Results for this expanded case set are shown in Figure~\ref{fig:expanded}.

Figures~\ref{fig:original} and \ref{fig:expanded} show receiver-operating characteristic (ROC) curves for GPT-5-mini with high reasoning, representing the model’s ability to discriminate between patients who do and do not require emergency care. For comparison, the figures also show results for GPT-5.4 high reasoning, DeepSeek V4 Pro high reasoning and Claude Fable 5 high reasoning. In Extended Figure \ref{extfig:roc}, we further present the ROC curves for larger and smaller variants of each model family---GPT 5.4 and GPT-5-mini, Claude 5 Fable and Sonnet, and DeepSeek V4 Pro and Flash---evaluated at three reasoning levels.

\begin{figure}[htbp]
  \centering
  \includegraphics[width=0.75\linewidth]{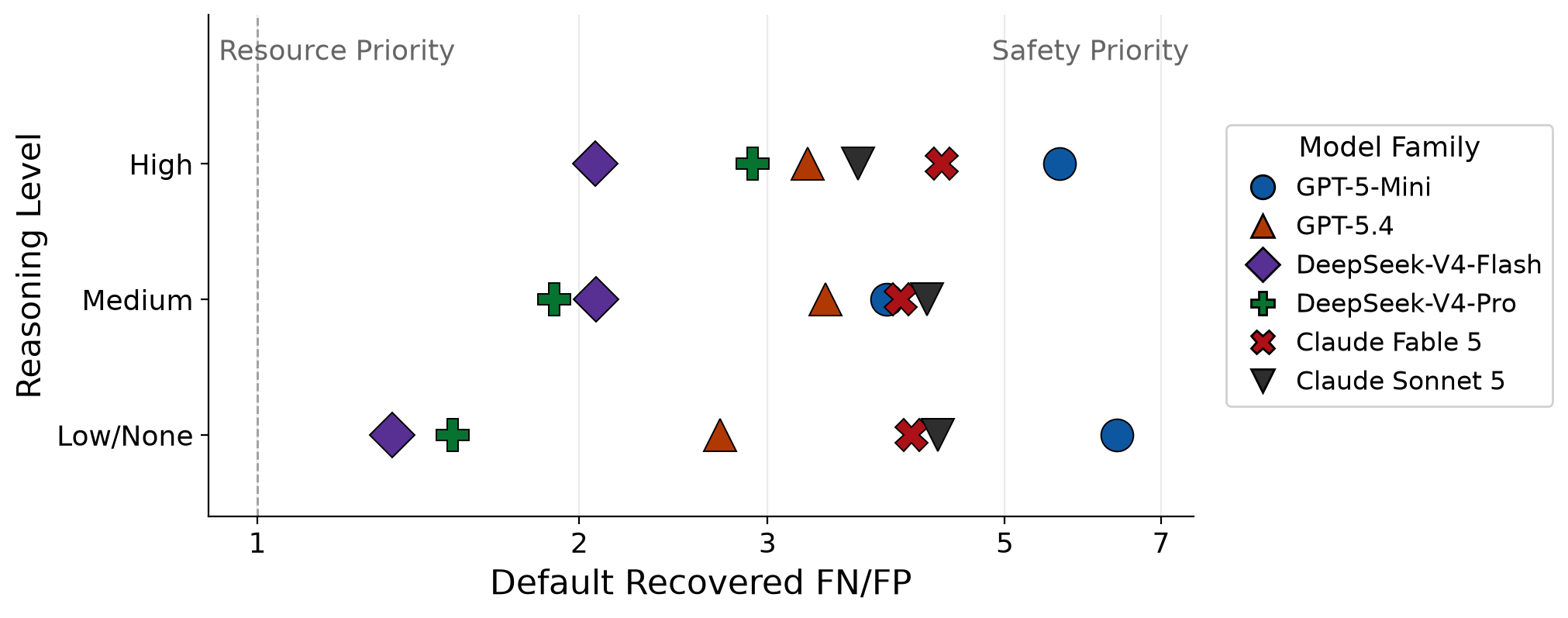}
  \caption{\textbf{Models' default utilities when none are specified.} Recovered default priorities for each model, grouped by model family and reasoning level. The defaults vary widely across models and shift with reasoning level, indicating that relying on an unstated model default means accepting an uncontrolled, model-dependent trade-off. Because model families use different labels for reasoning intensity, the categories shown in the plot are defined relative to the reasoning levels evaluated within each family. DeepSeek offers No Reasoning, High, and Max; GPT models offer Minimal/None, Medium, and High; and Claude models offer Low, Medium, and High.}
  \label{fig:defaults}
\end{figure}

Each point on the ROC curves in Figures~\ref{fig:original} and \ref{fig:expanded} corresponds to a different probability threshold for referral: moving up and to the right catches more true emergencies but increases unnecessary referrals, whereas moving down and to the left reduces unnecessary referrals while missing more emergencies. The highlighted points show where the deployed tool and the same underlying models, when instructed with explicit utilities over outcomes, fall on this trade-off curve.

For the primary endpoint in Figure~\ref{fig:original}, which includes as emergencies only cases in which adjudicating physicians unambiguously believed that urgent emergency-department care was required, we found that the GPT Health deployed tool's default behavior, generated without explicit instruction about priorities, sits near the extreme resource-prioritized end of this curve. It rarely refers patients unnecessarily, with a false-alarm rate of about 10\%, but catches only about 48\% of true emergencies. The other model families shown have different default behavior, catching 94--100\% of true emergencies with false-alarm rates ranging from 12\% to 28\%. However, AUROCs for the different models lie in the range 0.95--0.99, indicating that all models we evaluate are highly capable at ranking which cases are more or less likely to require emergency care. We therefore hypothesize that performance differences are attributable mostly to the implicit preferences that each model instantiates, not to their diagnostic capability.

To investigate this possibility and test whether implicit preferences can be steered in a desired direction, the colored points on Figure~\ref{fig:original} show the results of prompting each model with an explicit cost ratio between missed emergencies and unnecessary referrals.  Prompting GPT-5-mini reasoning to prioritize safety, using cost ratios of five or ten for missed emergencies relative to unnecessary referrals, moves the operating point up the trade-off curve: the share of true emergencies correctly routed increases by roughly 50$\%$ without increasing unnecessary referrals. Across all models, the decisions induced by utility prompts span most of the ROC curve, reaching roughly 100\% sensitivity. Much of the reported under-triage can therefore be explained by an implicit preference against over-referral rather than by limited discriminatory performance alone. 

The same pattern holds for the extended case endpoint. Here, AUROCs remain high and span between 0.88 and 0.94 for the various models on their high-reasoning mode.  Further, the colored points in Figure \ref{fig:expanded} show that prompting the same model behind ChatGPT Health
(GPT-5-mini reasoning) to prioritize safety, using cost ratios of five or ten, moves the operating point up the trade-off curve. The share of
true emergencies correctly routed increases by roughly 30$\%$, while unnecessary referrals increase only marginally. For other models, both those shown in  Figure \ref{fig:expanded} and the wider set shown in Extended Figure \ref{extfig:roc}, we similarly find that prompting with explicit cost ratios moves behavior in the desired direction along the trade-off curve, with most of the curve accessible via prompting for highly capable models.

Two caveats are important. First, the inferred default threshold does not fully explain the performance gap observed for GPT-5-mini, the model reportedly used in ChatGPT Health. As shown in Figure \ref{fig:original}, even at the same false-alarm rate, the tool identifies fewer emergencies than would be expected on the basis of the model's probability estimates (though the discrepancy is significantly lower on the expanded case set). One possible explanation is that the deployed tool does not act according to a single, internally consistent utility function. For stronger models we evaluate, default performance lies closer to the Pareto frontier achievable from their probability estimates, indicating a greater degree of internal consistency.

Second, we cannot exactly reproduce ChatGPT Health. The version of the deployed service studied in Ramaswamy et al.~(2026)\cite{ramaswamy2026chatgpthealth}  is no longer publicly accessible and its system prompt, safeguards, and exact configuration have not been disclosed. Our goal is therefore not to audit the commercial health service,  but to use a decision-analytic perspective to gain insight into the role of implicit and explicit priorities. When priorities are left unstated, the model can be viewed operating according to an implicit default; when priorities are stated, capable models can be steered toward the requested operating point.

Importantly, the model's default priorities could not be easily anticipated by clinicians. Using all 1,248 case variants, Figure~\ref{fig:defaults} shows the priorities implied when none  are explicitly stated, expressed as the relative weighting of false negatives and false positives and inferred from the behavior of different variants within the GPT, Claude, and DeepSeek model families. Because these analyses compare each model's decisions with its own elicited probabilities rather than with external clinical labels, every case provides information about the model's implicit preferences. The larger set of 1248 variants therefore enables more precise estimation of those preferences. The default priorities vary substantially across model and reasoning settings, ranging from moderate to strong safety weighting. Deploying a model with no stated priority therefore means accepting an implicit value judgment that is both unobservable in advance and variable.

\section*{Decision-analytic steering}
\begin{figure}[htbp]
  \centering
  \includegraphics[width=\linewidth]{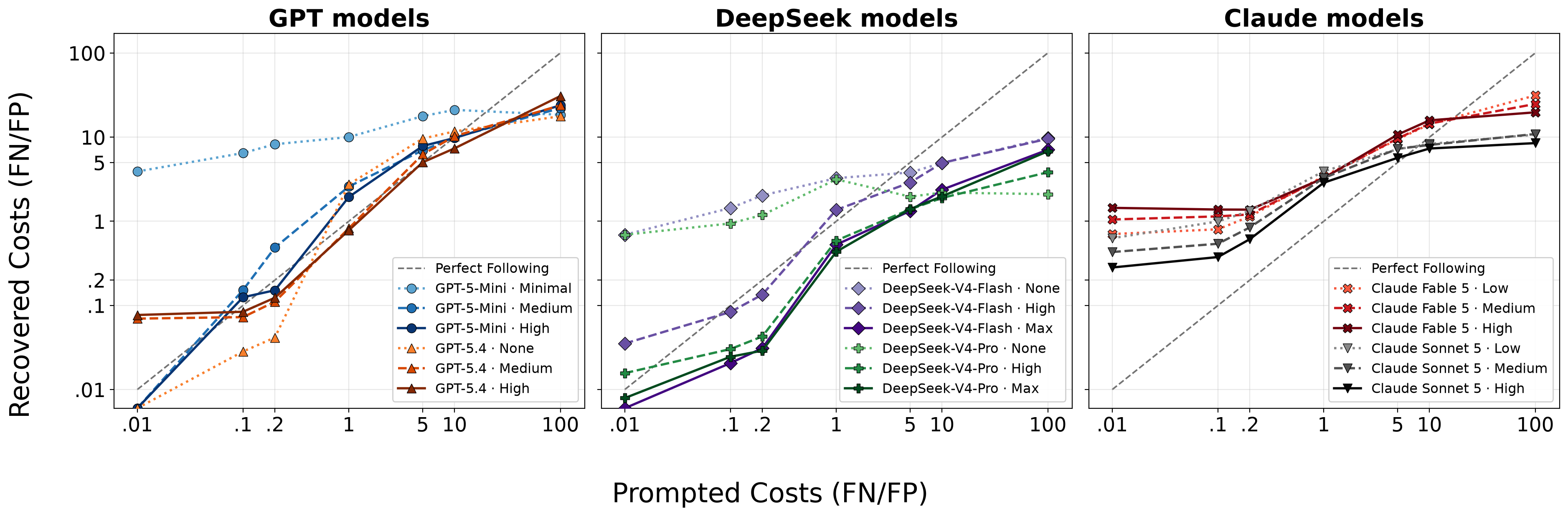}
  \caption{\textbf{Capable models adopt the priorities they are prompted with.}
  Each line is one model configuration. The horizontal axis is the cost ratio between a missed emergency and an unnecessary referral; the vertical axis shows the cost ratio recovered from the model's decisions. The dashed line marks perfect agreement. Hue denotes model family, while shading and line style denote reasoning level: light and dotted lines indicate minimal or no reasoning, and dark, solid lines indicate high reasoning. For the GPT and DeepSeek model families, more
  capable configurations track the instruction closely across the full range; the
  exceptions are the smallest models run without reasoning (light, dotted), whose
  recovered priorities remain high and largely unmoved regardless of instruction. For Claude, all models react to steering by shifting their decision rule in the correct direction, but the magnitude of the shift is generally dampened.}
  \label{fig:recovered}
\end{figure}

Using the same full set of 1,248 variants, we next examine how faithfully models follow prompted triage priorities, a capability we term decision-analytic steering. Figure~\ref{fig:recovered} compares each prompted cost ratio with the  corresponding ratio recovered from the model’s decisions. Across model families, such prompting generally shifts behavior in the intended direction: models recommend emergency care more often when instructed to prioritize avoiding missed emergencies and do so less often when instructed to avoid unnecessary referrals. 

The strength of decision-analytic steering varies by model family and capability. GPT and DeepSeek models with inference-time reasoning enabled track the requested utilities closely across most of the range from resource-prioritized to safety-prioritized settings. Notably, GPT-5.4 follows the requested utilities even with minimal reasoning, whereas GPT-5-mini and DeepSeek models with reasoning disabled remain confined to a narrow range of thresholds despite large changes in the prompt. Claude Fable 5 and Claude Sonnet 5 consistently shift their decisions in the intended direction across reasoning levels, but their responses are dampened, with recovered cost ratios spanning a narrower range than the prompted ratios, especially at the extremes (e.g., where false negatives are weighted 0.1 or 100 times a false positive).  Overall, reliable steering appears to depend on either sufficient underlying model capability or explicit inference-time reasoning, and even capable models differ in how faithfully they implement the requested trade-off. An alternative prompting scheme that directly specifies the probability threshold implied by each cost ratio produces similar results, as shown in Extended Figure~\ref{extfig:recoveredthreshold}. We conclude that decision-analytic steering is a useful tool, particularly for more capable models, but not one whose success can be assumed.

\section*{Calibration and decision thresholds} An alternative to utility prompting is to derive a fixed decision rule from expert-reviewed cases, such as those in the dataset from Ramaswamy et al.~(2026) analyzed here. For each model, the colored squares in Extended Data Figure~\ref{extfig:roc} show the optimal fixed-threshold operating point under each prespecified false-negative-to-false-positive cost ratio. The target ratio represents the clinicians' desired trade-off; for example, a ratio of five assigns five times as much cost to a missed emergency as to an unnecessary referral. For each target ratio, we select the model-specific probability threshold that minimizes the corresponding cost-weighted error. Because models may be differently calibrated, the threshold needed to realize the same target trade-off can vary across models. Expressing the selected threshold in utility terms yields an equivalent model-specific cost ratio, which we refer to as the best fixed utility ratio. Thus, the target ratio states the desired clinical priority, whereas the best fixed utility ratio describes the model-specific decision rule that most closely realizes that priority. For example, the yellow square in Figure~\ref{fig:original} corresponds to  a target ratio of five, which prioritizes safety; its location on the ROC curve identifies the threshold, and therefore the equivalent best fixed utility ratio, that minimizes cost-weighted error for that model. 

This analysis suggests a practical pathway for utility-guided decision support in high-stakes settings. If clinicians at a hospital provide a modest sample of locally reviewed cases and specify a target cost ratio, the best fixed utility ratio for implementing their desired trade-off can be recovered automatically. This ratio defines an explicit operating point on the safety--resource trade-off, rather than leaving the model to rely on an unstated default. For subsequent cases, the corresponding utility specification can either be conveyed in the prompt or implemented externally as a fixed threshold applied to the model's elicited probabilities. Either approach can support decision-analytic alignment by bringing recommendations into closer agreement with local clinical priorities. 

Externally selecting a threshold also provides some robustness to miscalibration. Although all models evaluated here discriminate well between lower- and higher-risk patients, Extended Data Figure~\ref{extfig:beliefdist} and \ref{extfig:calibration} show that most models produce poorly calibrated probabilities. Selecting a threshold empirically from reviewed cases can compensate for systematic miscalibration by locating the decision boundary that best implements the target trade-off. This approach does not recalibrate the probabilities themselves, but it can preserve decision performance as long as the models' risk rankings remain informative and the relationship between estimated risk and observed outcomes remains stable over time, or is periodically re-estimated.

\section*{Discussion}

These results show that the central problem in harnessing language models for emergency triage is not only whether a model recognizes urgency, but also whether it applies the appropriate decision policy to translate that assessment into action. This distinction motivates a decision-analytic paradigm for the steering, evaluation, and deployment of language models in consequential settings. In our experiments, capable models all ranked patients well. The deployed tool's unsafe default behavior was not simply a failure of medical knowledge; it was substantially attributable to the decision policy applied to the model's assessments, including the absence of an explicitly specified utility. A model's default behavior is not neutral. It implicitly assigns relative costs to different errors, with no reason to expect the resulting priorities to align with those of users, domain experts, institutions, or affected stakeholders. Moreover, these implicit utilities  vary across models and reasoning settings. Treating default behavior as inherently appropriate or normatively meaningful without explicit specification is therefore unsafe and inconsistent with decision-analytic alignment.

We draw three recommendations for language models used in high-stakes decision support. First, prompts should state the intended priorities in ordinary language, specifying how the costs of different errors should be weighed. This is analogous to setting an operating threshold in a clinical protocol. Our results show that capable models consistently respond to such instruction. However, they do not always respond in a fully precise fashion, and the we suggest improving the precision with which models implement stated priorities as an important direction for model developers. Second, models should be evaluated across a range of priorities rather than at a single default operating point. A single performance estimate may make a steerable model appear unsafe because it is evaluated under an inappropriate implicit trade-off, or make a poorly controlled model appear acceptable merely because its default matches the evaluator's preference. Evaluations should instead report the full trade-off curve and test whether the model reliably reaches the requested operating points.\cite{cruz2024evaluating,burnell2023rethink} Third, systems should be provided with or elicit explicit priorities before recommending actions, whether decision-analytic alignment is pursued through prompting, fine-tuning, external thresholds, or hybrid system designs. When the intended trade-off is unspecified, a system should seek clarification or apply a clearly stated safety-prioritized default. Just as high-stakes systems must pursue clarification of uncertain situations or facts, they must also clarify the values governing action.\cite{li2024mediq}

\section*{Methods}

\paragraph{Overview and framing.} Our methods provide a decision-analytic perspective on the use of language models. We treat each triage decision as combining two
distinct ingredients: a probabilistic assessment of whether a patient truly needs
emergency care, and a \emph{value} judgment (encoded as a cost ratio) about how
to weigh a missed emergency against an unnecessary referral. Our procedure elicits
these two ingredients separately and then relates them, building on the
revealed-preference framework of Yamin and colleagues\cite{yamin2026revealedpreferencesclarifyllm}.
The entire pipeline is black-box: it uses only prompts and the models' text
responses, with no access to model internals. The utility-prompting component operationalizes decision-analytic steering, whereas the external-threshold analysis illustrates an alternative route to decision-analytic alignment.

\paragraph{Clinical material.} We reused the clinician-authored triage vignettes
from the structured stress test of Ramaswamy and
colleagues,\cite{ramaswamy2026chatgpthealth} expanded across the study's 16
factorial conditions, which varied patient race, sex, presence of an anchoring
cue and stated barriers to care. Because the central safety concern in the
original study was failure to escalate a genuine emergency, we reduced the
clinician-adjudicated urgency assessments to a binary endpoint---whether the patient
requires urgent emergency-department care---classifying a variant as a true
emergency when the adjudicating physicians unambiguously believed that urgent
emergency-department care was required. Following the methodology of the original
study's primary emergency analysis, we restricted attention to its 960 original
vignette variants and excluded cases in which physicians believed that the patient
needed urgent attention, either from a physician or in an emergency department. This yields
576 case variants, of which 64 are true emergencies and 512 are non-emergencies.
Figure~\ref{fig:expanded} instead uses the expanded set of 1,248 variants, which
includes supplementary textbook emergencies and treats the excluded cases as
emergencies; this yields 640 emergencies and 608
non-emergencies. All utility-recovery and decision-analytic-steering analyses,
including Figures~\ref{fig:defaults} and \ref{fig:recovered}, use the full 1,248
variants: these analyses compare each model's decisions with its own elicited
probabilities rather than with external clinical labels, and the larger sample
better estimates the model's preferences.
Eliciting probability estimates and decisions from the identical context for each
variant enables a direct comparison between them.

\paragraph{Models.} We evaluated GPT-5-mini,GPT-5.4, Claude Fable 5, Claude Sonnet 5, DeepSeek V4 Pro and DeepSeek V4 Flash each at 3
test-time reasoning settings (spanning from no reasoning to high reasoning) yielding  18
model configurations in total. The consumer tool that motivated the study was
documented as running on a small GPT-5 model with a thinking or reasoning setting; we
therefore include the same small model family for the direct comparison in
Figure~\ref{fig:original}, because the product itself is no longer broadly
accessible for querying and its exact system prompt and reasoning configuration
are not public.

\paragraph{Eliciting probabilities.} For every case variant, we asked the model, in an
isolated query, to report the probability that the patient requires emergency
care given the clinical description. The question was deliberately plain, with no
mention of costs, thresholds or subsequent actions, so that the reported
probability would reflect the model's factual assessment rather than a response strategically shaped by an anticipated action. We treat this probability as the model's expressed probabilistic belief and use it as the common yardstick against which its decisions are
interpreted. Probability estimates were elicited in a separate context from the decision
queries so that the two could not directly contaminate one another, an approach
validated in prior work using the same elicitation and shown there to be stable
under repeated sampling and small perturbations.\cite{yamin2026revealedpreferencesclarifyllm,paruchuri2024odds}

\paragraph{Eliciting decisions and stating priorities.} In separate queries, we
asked the model to provide a triage recommendation. We used 8 prompting regimes:
a \emph{baseline} regime with no stated priorities 
and seven \emph{priority} regimes in which we appended, in plain language, a
specific cost ratio for the two error types. The seven ratios spanned strongly
resource-prioritized to strongly safety-prioritized (missed-emergency-to-unnecessary-referral
costs of roughly 1:100, 1:10, 1:5, 1:1, 5:1, 10:1 and 100:1). The baseline regime approximates
how a user or evaluator might interact with a triage tool out of the box, without
supplying explicit utilities. In the priority regimes, the cost ratio was conveyed in
ordinary language. For example, we told the model that failing to send a patient
who truly needs emergency care is a certain number of times more, or less, serious
than sending a patient who does not need emergency care. This phrasing was designed to resemble guidance that a clinical service might provide rather than a mathematical formula. Each
regime was applied to all case variants, and the probability-estimation and decision
queries were kept in separate contexts.

\paragraph{Recovering the cost ratio implied by a model's behavior.} Given a
model's probability estimates and decisions, we recovered the single cost ratio that
best explained those decisions. Intuitively, if a model refers every patient whose
estimated emergency probability exceeds some threshold, that threshold pins down
the ratio of the two error costs that best describes its behavior. A low threshold implies a
safety-prioritized ratio whereas a high threshold implies a resource-prioritized one. Under standard decision
theory, a cost-minimizing decision maker should refer a patient whenever the
probability of a true emergency exceeds a break-even value determined entirely by the cost ratio. This threshold equals the false-positive cost
divided by the sum of the false-positive and false-negative costs. A ten-to-one safety-prioritized ratio
therefore implies referral whenever the probability of an emergency exceeds approximately one in 
eleven, while a ten-to-one resource-prioritized ratio implies referral only when the probability exceeds approximately ten in eleven. 

To recover a model's utilities, we estimated the single
break-even probability, and hence the single cost ratio, whose implied referrals
best reproduced the model’s actual binary decisions across all cases, given its own elicited probabilities. We performed this analysis separately for each model and 
prompting regime. Formally, we fit a
standard discrete-choice logistic model relating decisions to elicited probabilities and derived the implied cost ratio, following established practice in choice
modeling.\cite{mcfadden1973conditional,train2009discrete} Only the \emph{ratio}
of costs is identifiable and reported; their absolute scale is not.  We therefore report cost ratios rather than separate costs. The recovered
ratio is plotted on the vertical axis of Figure~\ref{fig:recovered} and used
to summarize each model's default, no-priority stance. We validated the recovered
ratios in two ways: they moved in step with the explicitly prompted ratios in Figure~\ref{fig:recovered}, and the corresponding decision rules reproduced most of the models' actual choices in Extended Data Figure~1.

\paragraph{Assessing discrimination and the trade-off curve.} To assess how well a
model's probability estimates distinguish true emergencies independently of any value judgment, we
compared the elicited probabilities with the gold-standard labels using
receiver-operating-characteristic analysis and reported the area under the curve as a summary of ranking performance in Figure \ref{fig:original} and Extended Figure \ref{extfig:roc}. Because each cost
ratio corresponds to a probability threshold, and each threshold corresponds to an
operating point on the curve, the same analysis allows us to overlay the decisions made
under each prompted ratio and test whether the prompted utility moves the model along the curve as intended.

\paragraph{Self-consistency.} As an internal check, we measured how often each
model's decisions agreed with the rule ``refer whenever the
model's own estimated probability exceeds the threshold implied by its recovered
cost ratio.'' High agreement indicates that a model's behavior is well described
by the combination of its expressed probability estimates and a single utility setting, rather than by an
unstable or inconsistent decision rule. These results are reported in Extended Data Figure \ref{extfig:consistency}.

\paragraph{Comparison with deployed consumer health tool.} For Figure~\ref{fig:original}, we
placed the triage behavior of ChatGPT Health as reported in Ramaswamy et al. (2026),\cite{ramaswamy2026chatgpthealth} summarized by its sensitivity and unnecessary-referral rate, on the trade-off curve implied by the probability estimates of the same small model family that the tool was documented to use. We assessed both where the tool's default lay along the
resource-prioritized portion of the curve and whether it fell below
the curve at its observed unnecessary-referral rate. This analysis distinguishes the portion of the
under-triage that can be explained by the operating threshold, and is therefore potentially movable by prompting, from the portion associated with a discrimination gap that cannot be corrected by any single cost ratio. As a consistency check on the choice of reference model, we
also measured agreement between our elicited decisions and the deployed tool's published decisions across 
prompting regimes. Agreement was highest for the small model family that the tool
was documented to use (Extended Data Figure \ref{extfig:natureconsistency}).

\paragraph{Statistical reporting.} Where uncertainty is shown, intervals are 95\%
confidence intervals obtained by bootstrap resampling over cases. Because the parsing
of free-text responses occasionally failed, sample sizes varied slightly across regimes;
these differences did not materially affect the reported patterns. We focus on effect sizes and qualitative patterns rather than detailed significance testing because our primary questions concern
whether models follow stated priorities, whether their 
 probability estimates discriminate between cases, and where their default behavior lies on the trade-off curve.

\paragraph{Ethics and data.} The study used clinician-authored synthetic vignettes
and publicly reported model outputs. It involved no human subjects and no
identifiable patient data.

\paragraph{Data Availability}
The original triage cases are available at: \url{https://zenodo.org/records/18451491} \cite{ramaswamy2026chatgpthealth}. The datasets we generate that include the LM's inferred probability on cases, and decisions made by the LMs under various prompting regimes are available at: \url{https://github.com/khurramyamin/LLM-Triage-Experiment/tree/main/data}. We release one file per model, covering every reasoning-effort setting and both prompting regimes (stated misclassification costs and the equivalent probability thresholds): each row records the elicited probability that a patient needs emergency care or the decision made, alongside the clinician-adjudicated gold standard, the case metadata, and the model's full response.

\paragraph{Code availability}
All code used for our project is made open-source and can be accessed via: \url{https://github.com/khurramyamin/LLM-Triage-Experiment}. Within this link, we further include code that can create ROC curves and generate the best fixed utilities for the data (\url{https://github.com/khurramyamin/LLM-Triage-Experiment/tree/main/generic_RoC}). This code can be used to analyze datasets capturing any domain involving probabilities and decision making.

\section*{Acknowledgments} 
We thank Xavier Fernandez and Paul Koch for their feedback.  


\clearpage
\bibliographystyle{unsrtnat}
\bibliography{references}

\section*{Supplementary Materials}

The supplementary materials include extended figures that present additional analyses of frontier model performance for the primary and expanded endpoints. The figures show results on discrimination, decision coherence, instruction following, and probabilistic calibration. Extended Data Figure~\ref{extfig:roc} presents ROC curves across model and reasoning configurations for the primary 576-case endpoint. Extended Data Figure~\ref{extfig:roc-expanded} shows the corresponding curves for the expanded 1,248-case endpoint. Extended Data Figure~\ref{extfig:consistency} assesses whether referral decisions are consistent with each model’s stated probabilities and recovered utility setting. Extended Data Figure~\ref{extfig:natureconsistency} compares the decisions of candidate models with those reported by ChatGPT Health,\cite{ramaswamy2026chatgpthealth} supporting the selection of the reference model. Extended Data Figure~\ref{extfig:recoveredthreshold} examines how closely models follow explicitly specified decision thresholds. These three behavior-based analyses use all 1,248 cases because they do not depend on the definition of clinical endpoint. Finally, 

Extended Data Figure~\ref{extfig:beliefdist} compares probability distributions and decision thresholds across configurations, highlighting the compressed ranges of probabilities of weaker models under both the 576- and 1248-case endpoints. Extended Data Figure~\ref{extfig:calibration} directly compares elicited probabilities with observed emergency-care frequencies and reports expected calibration error under both endpoints, using the clinical labels as reference outcomes. Extended Data Figure~\ref{extfig:prompts} presents the four structured prompts used to separately elicit  probabilities and decisions, and to specify decision priorities.

\begin{extendedfigure}[p]
    \centering
    \noindent\textbf{1a}
    \par\smallskip
    \includegraphics[width=\linewidth,height=0.50\textheight,keepaspectratio]{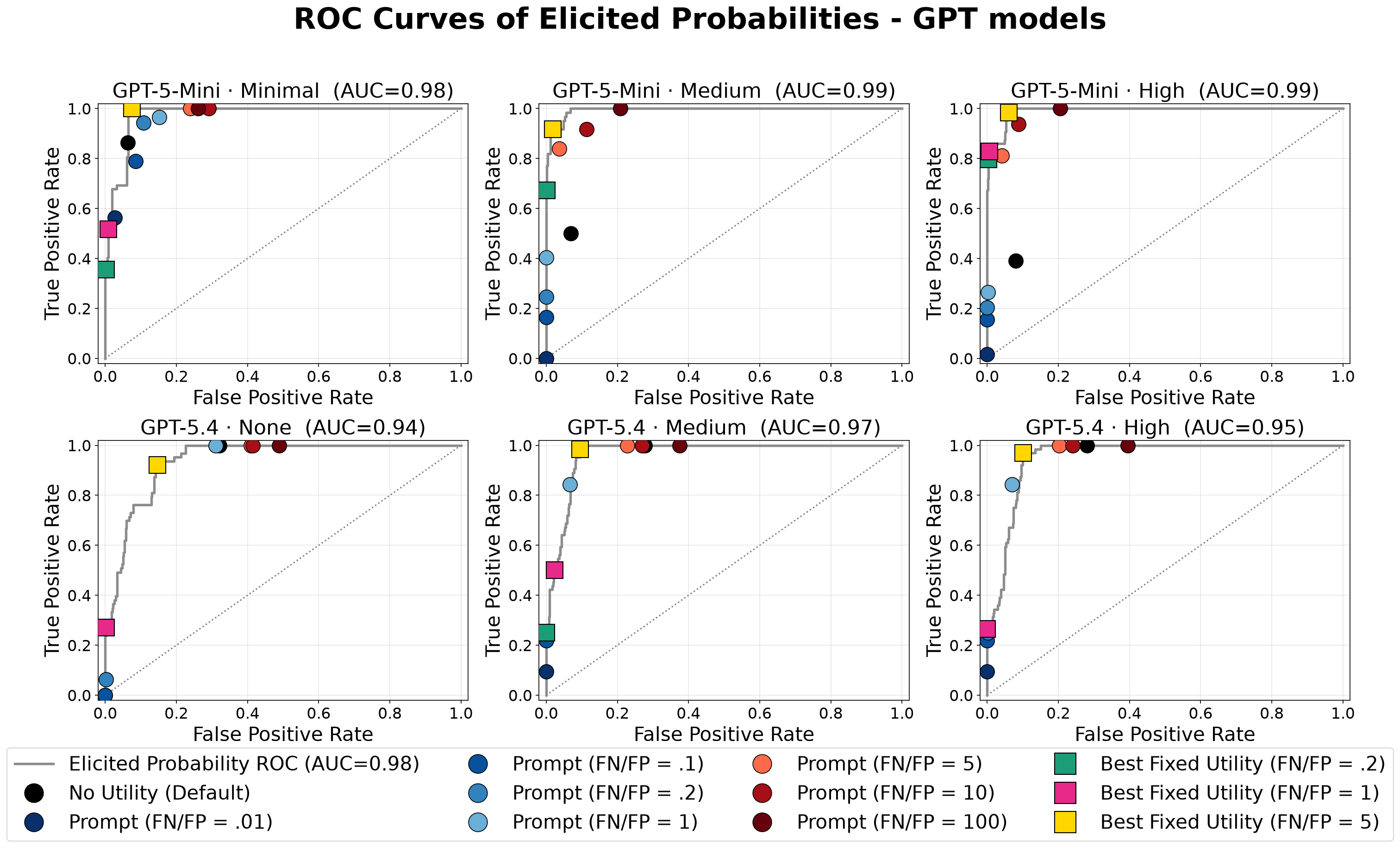}
    \caption{\textbf{ROC curves across model variants show strong probability rankings and utility trade-offs for emergency cases.} Results for (a) the GPT model family, (b) the DeepSeek model family and (c) the Claude model family with (b) and (c) on the following page. The ROC curve traces the trade-off between the proportion of true emergencies correctly referred to the emergency department (vertical axis) and the proportion of patients who do not require emergency care but are referred unnecessarily (horizontal axis), as the decision threshold is varied over the model's elicited probability estimates. A curve that bows toward the top-left indicates good discrimination; AUROCs, shown in each panel, range from 0.95 to 1. Colored points show the operating points produced by different prompted cost ratios, from strongly resource-prioritized (dark, lower-left) to strongly safety-prioritized (red, upper-right). The colored squares mark the optimal fixed-threshold operating points under three false-negative--to--false-positive cost ratios (green FN/FP = 0.2, pink FN/FP = 1, and yellow FN/FP = 5). For each target ratio, we select the single probability threshold that minimizes the cost-weighted number of errors relative to the reference clinical labels, with each missed emergency weighted FN/FP times as heavily as an unnecessary referral. This identifies the best operating point reachable from the model's probabilities under that priority. As the target ratio increases, the optimal fixed-threshold operating point moves along the ROC curve from the resource-prioritized lower left toward the safety-prioritized upper right. The fixed threshold can therefore be selected to reflect a site's preferred trade-off.}
    \label{extfig:roc}
\end{extendedfigure}

\begin{extendedfigure}[p]
    \ContinuedFloat
    \centering
    \noindent\textbf{1b}
    \par\smallskip
    \includegraphics[width=\linewidth,height=0.40\textheight,keepaspectratio]{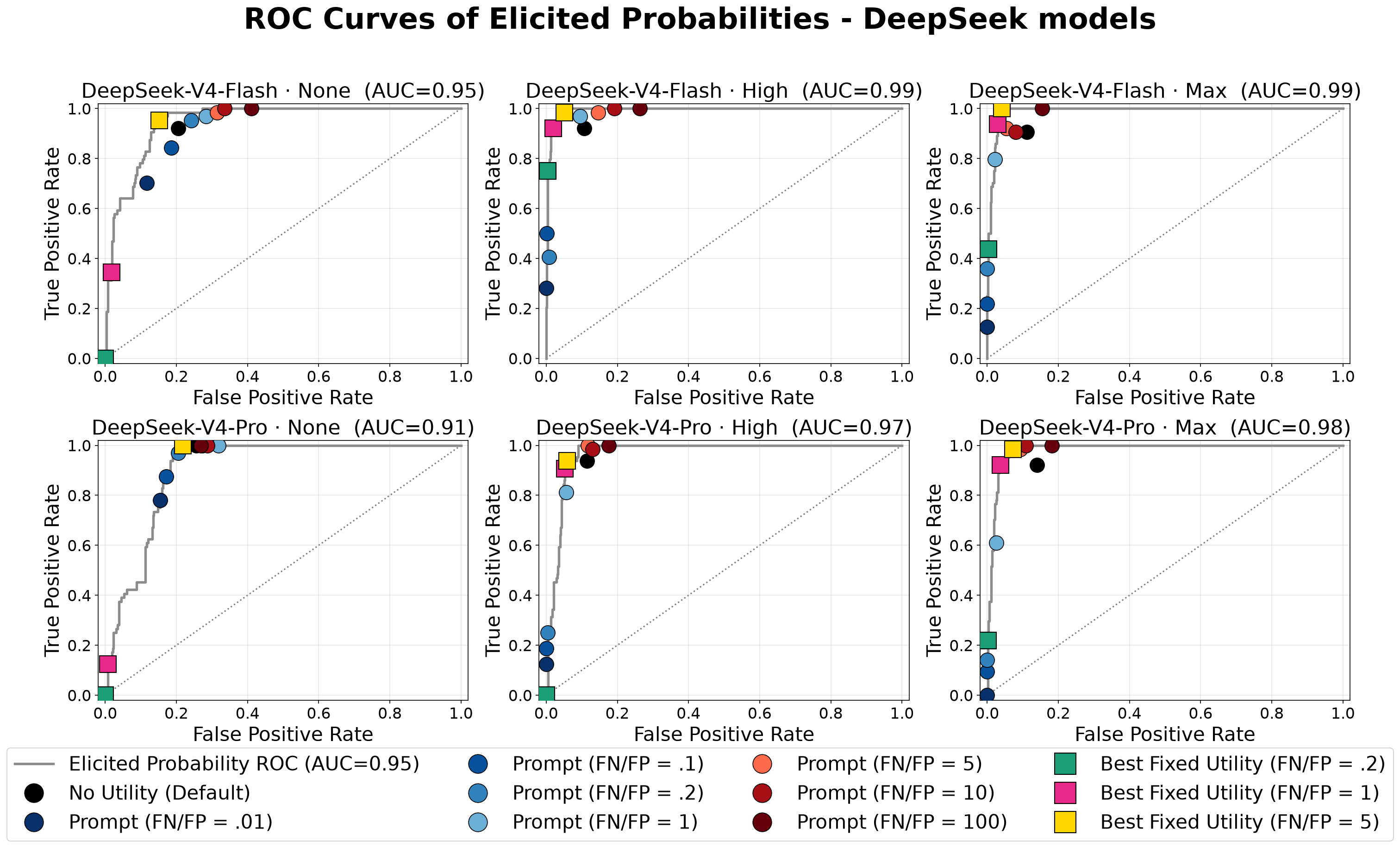}
    \par\smallskip
    \noindent\textbf{1c}
    \par\smallskip
    \includegraphics[width=\linewidth,height=0.40\textheight,keepaspectratio]{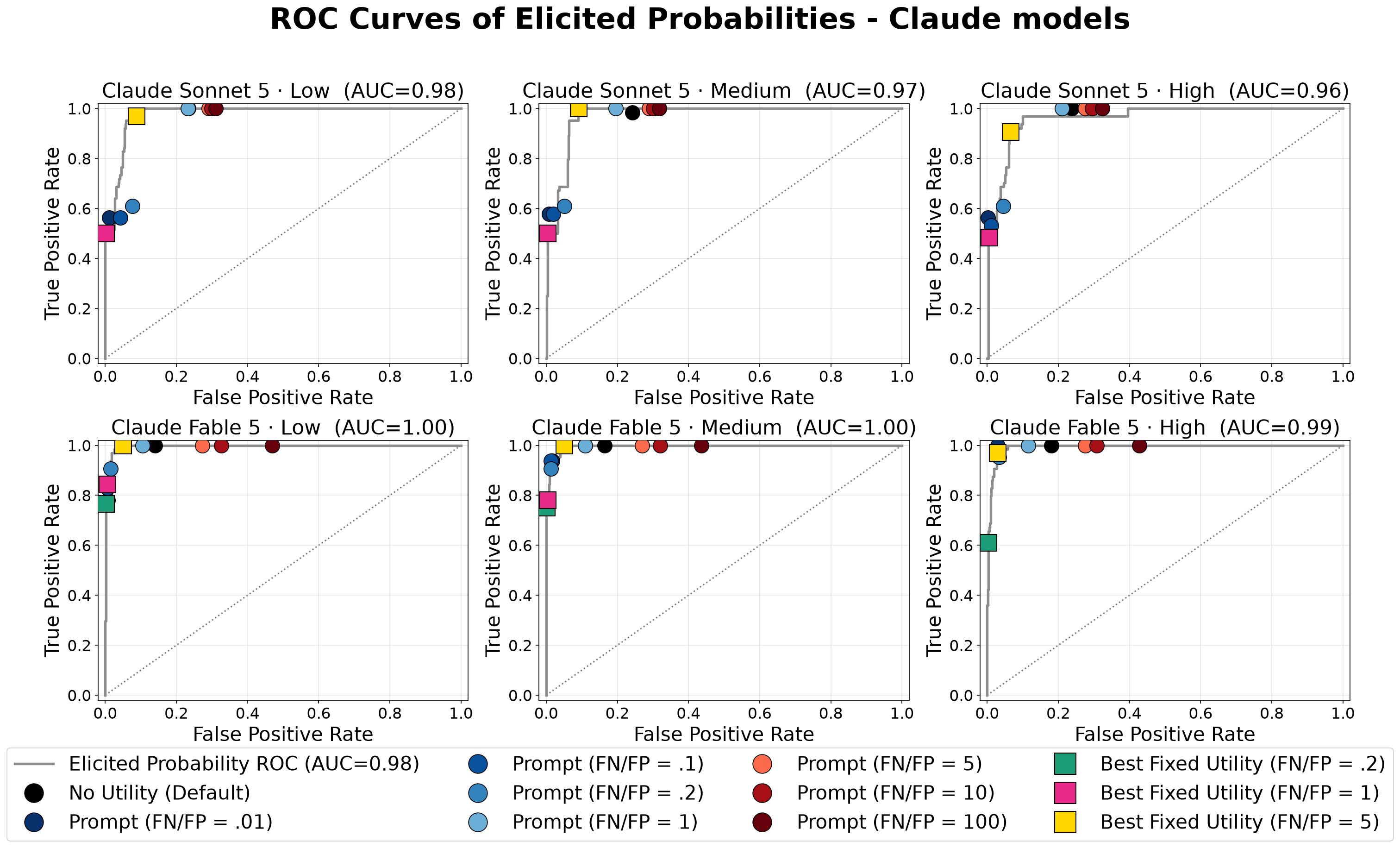}
\end{extendedfigure}

\begin{extendedfigure}[p]
    \centering
    \noindent\textbf{2a}
    \par\smallskip
    \includegraphics[width=\linewidth,height=0.50\textheight,keepaspectratio]{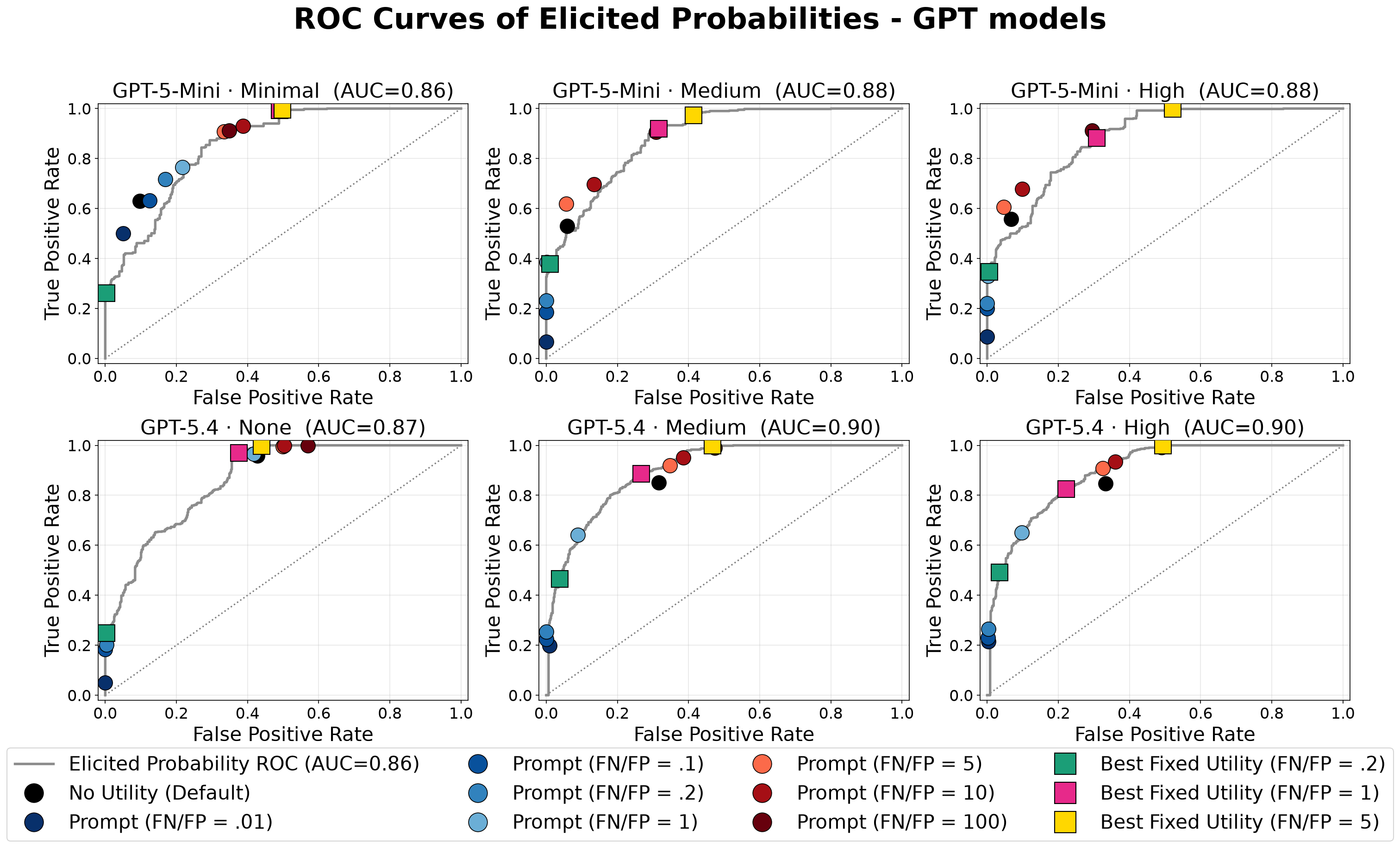}
    \caption{\textbf{ROC curves across models for the expanded endpoint.} Results for (a) the GPT model family, (b) the DeepSeek model family and (c) the Claude model family with (b) and (c) shown on the next page. The analysis includes all 1,248 cases and treats both definitive emergencies and ``edge'' cases as emergencies. Each panel shows one of 18 model configurations. ROC curves show the rankings implied by the model's elicited probabilities (AUROCs range from 0.86--0.94), circles show default and utility-prompted operating points, and squares show the optimal fixed-threshold operating points under false-negative--to--false-positive cost ratios of 0.2, 1 and 5. Other plot elements are defined in Extended Data Figure~\ref{extfig:roc}.}
    \label{extfig:roc-expanded}
\end{extendedfigure}

\begin{extendedfigure}[p]
    \ContinuedFloat
    \centering
    \noindent\textbf{2b}
    \par\smallskip
    \includegraphics[width=\linewidth,height=0.40\textheight,keepaspectratio]{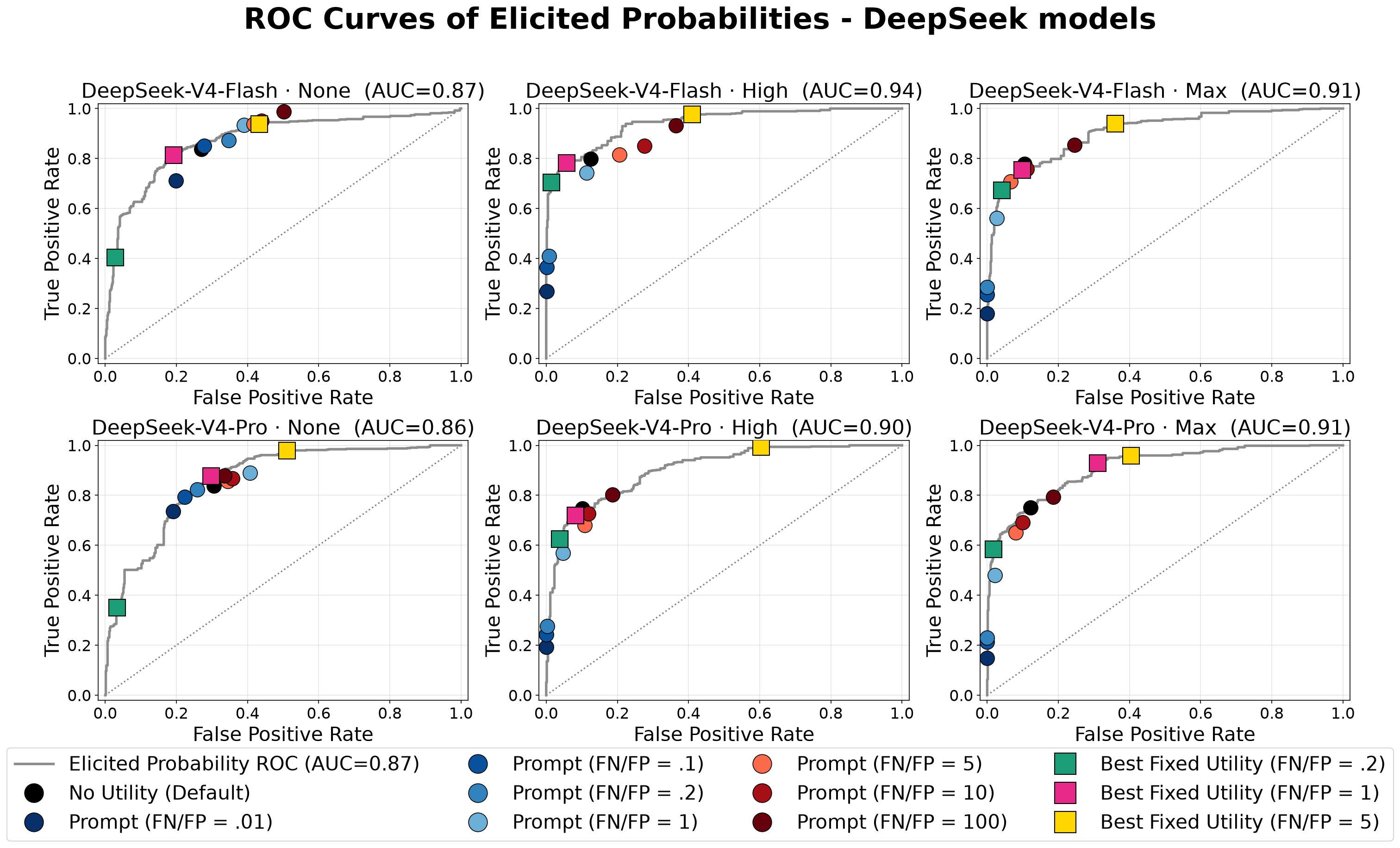}
    \par\smallskip
    \noindent\textbf{2c}
    \par\smallskip
    \includegraphics[width=\linewidth,height=0.40\textheight,keepaspectratio]{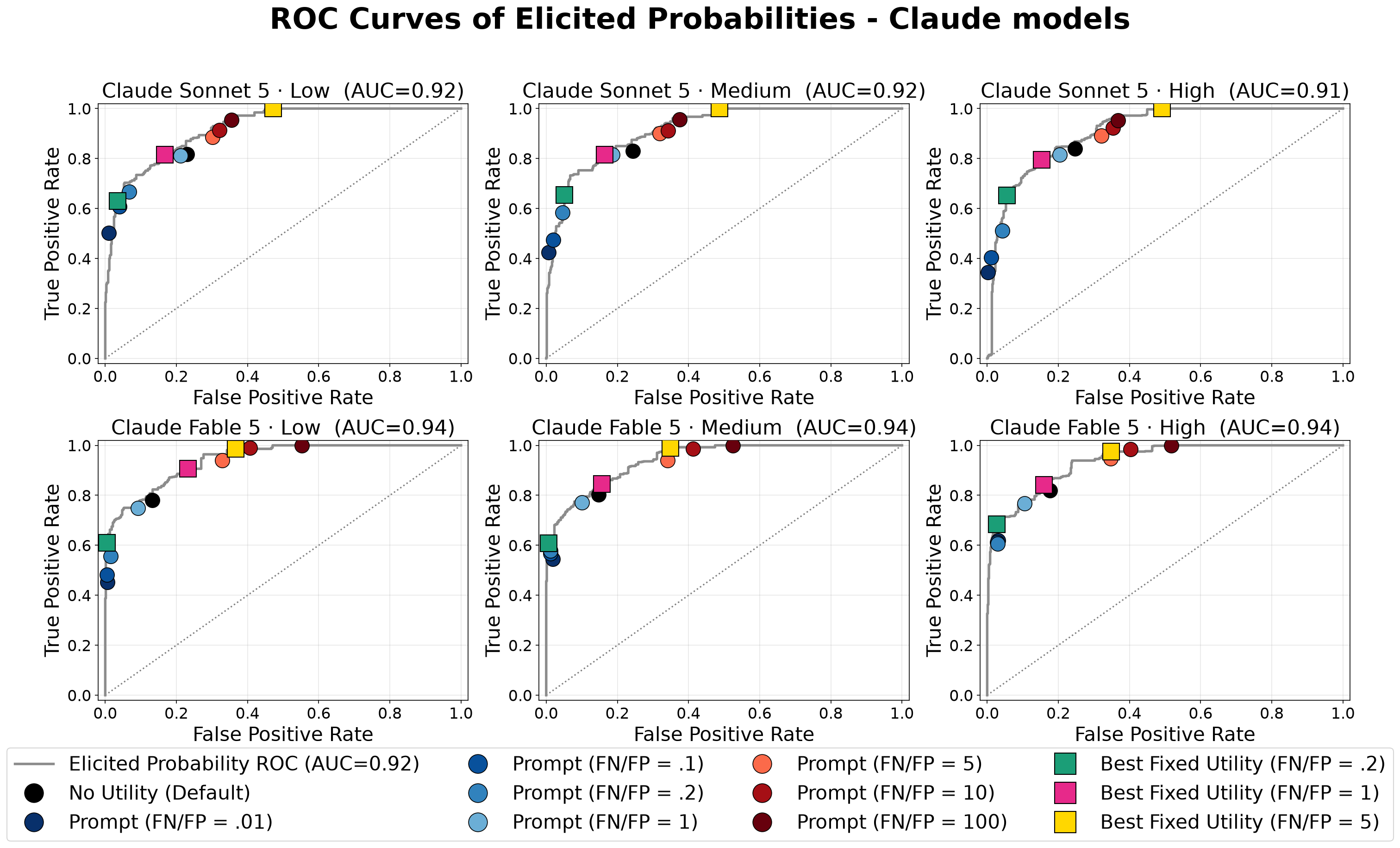}
\end{extendedfigure}

 \begin{extendedfigure}[htbp] \centering \includegraphics[width=0.66\linewidth]{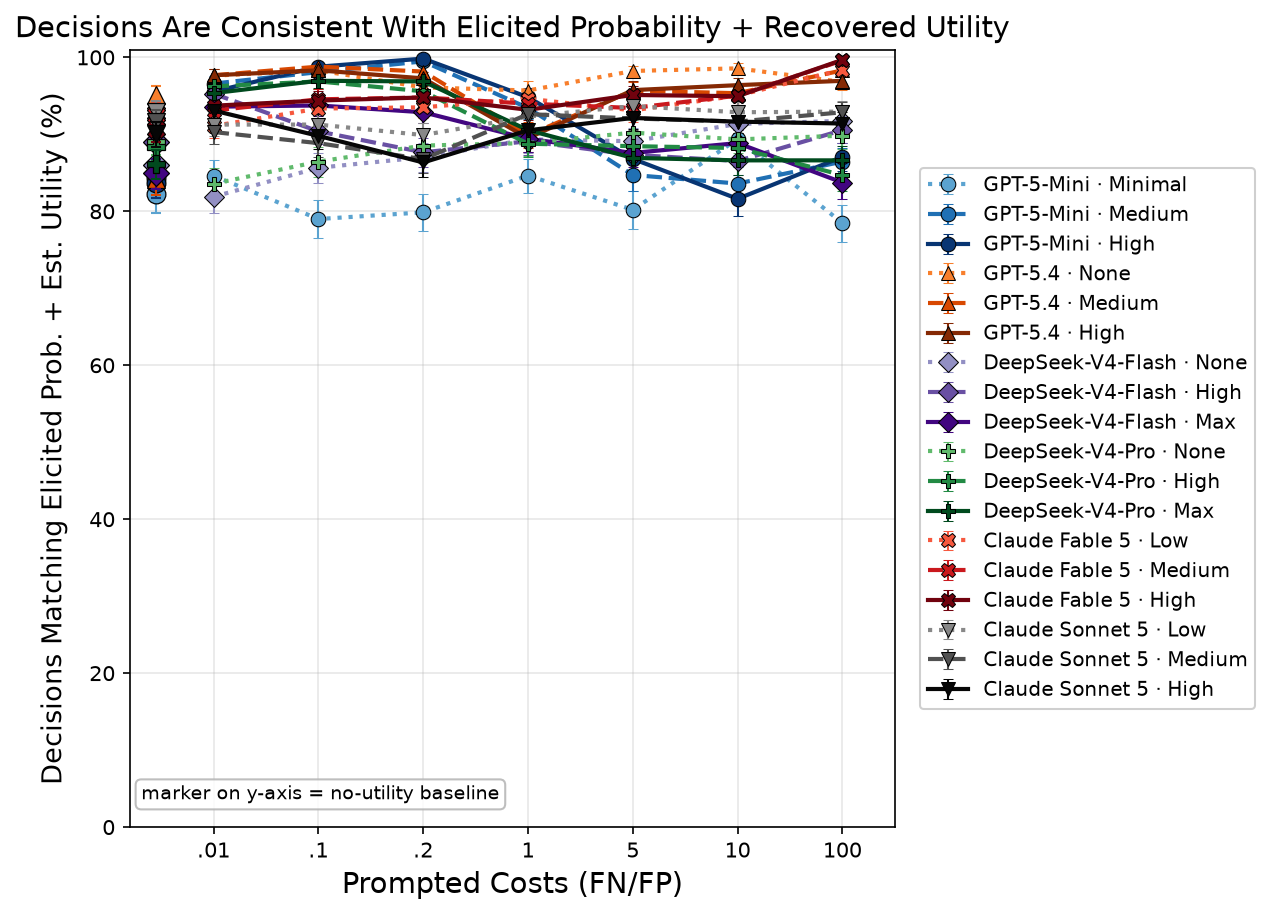} \caption{\textbf{Models generally make decisions consistently with their elicited probabilities.} For each model configuration and prompted cost ratio, the figure shows the proportion of decisions that follow the decision policy implied by the model's elicited probability and recovered cost ratio: ``refer to emergency care whenever the model's elicited probability exceeds the corresponding decision threshold.'' Agreement is high across models and prompting regimes, generally ranging from 80\% to 100\%. Thus, most decisions can be explained by the combination of the model's elicited probability estimate and a single recovered utility setting. Stars mark the default regime in which no priority is specified.} \label{extfig:consistency} \end{extendedfigure} \begin{extendedfigure}[htbp] \centering \includegraphics[width=0.66\linewidth]{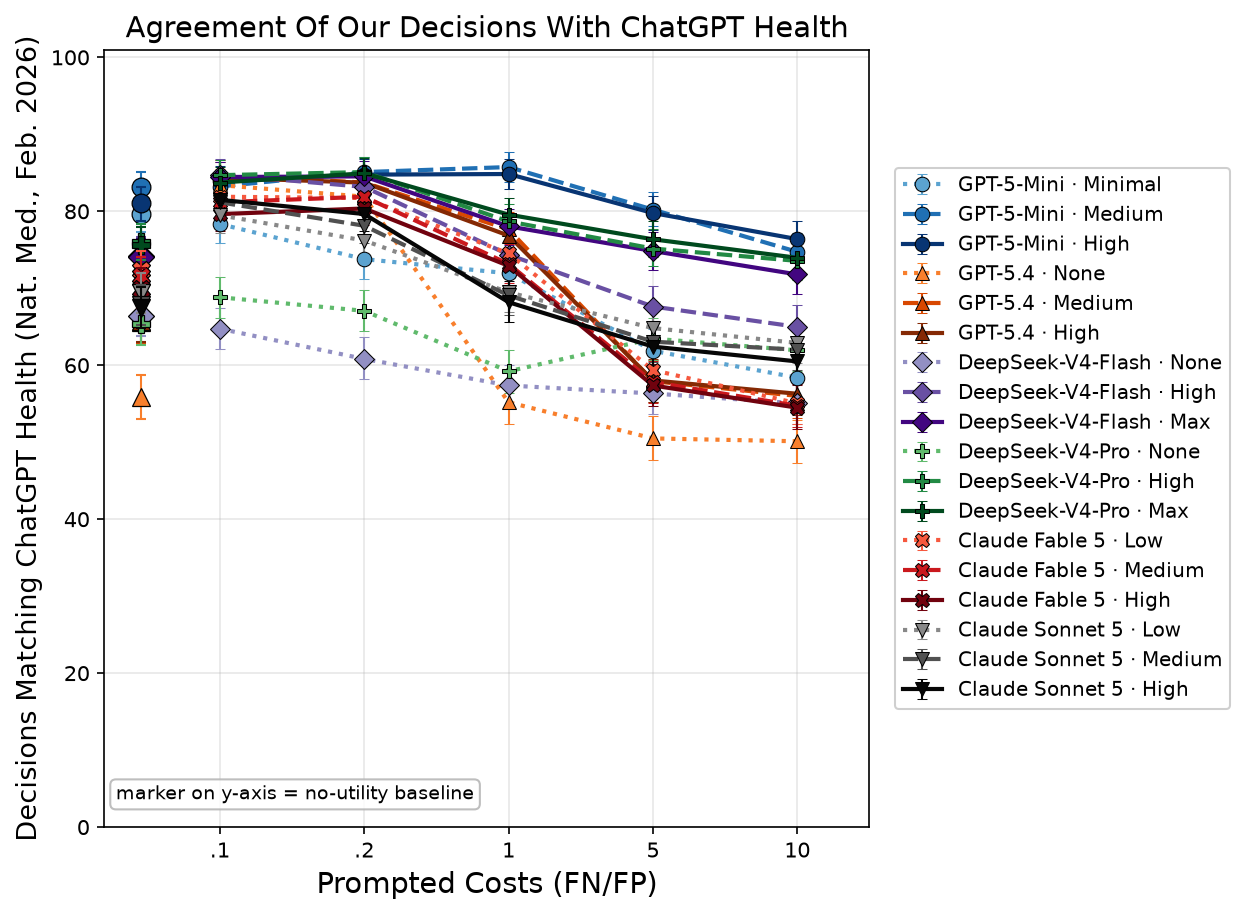} \caption{\textbf{The published decisions of ChatGPT Health most closely resemble those of GPT-5-mini.} The figure shows agreement between decisions elicited from each model configuration and the decisions published for ChatGPT Health, across prompted cost ratios. Agreement is highest for GPT-5-mini configurations with medium and high reasoning, consistent with reports that GPT-5-mini with reasoning served as the backbone of the version of ChatGPT Health studied by Ramaswamy et al.~(2026). This finding supports the choice of GPT-5-mini as the reference model for the comparison in Figure~\ref{fig:original}.} \label{extfig:natureconsistency} \end{extendedfigure} \begin{extendedfigure}[htbp] \centering \includegraphics[width=\linewidth]{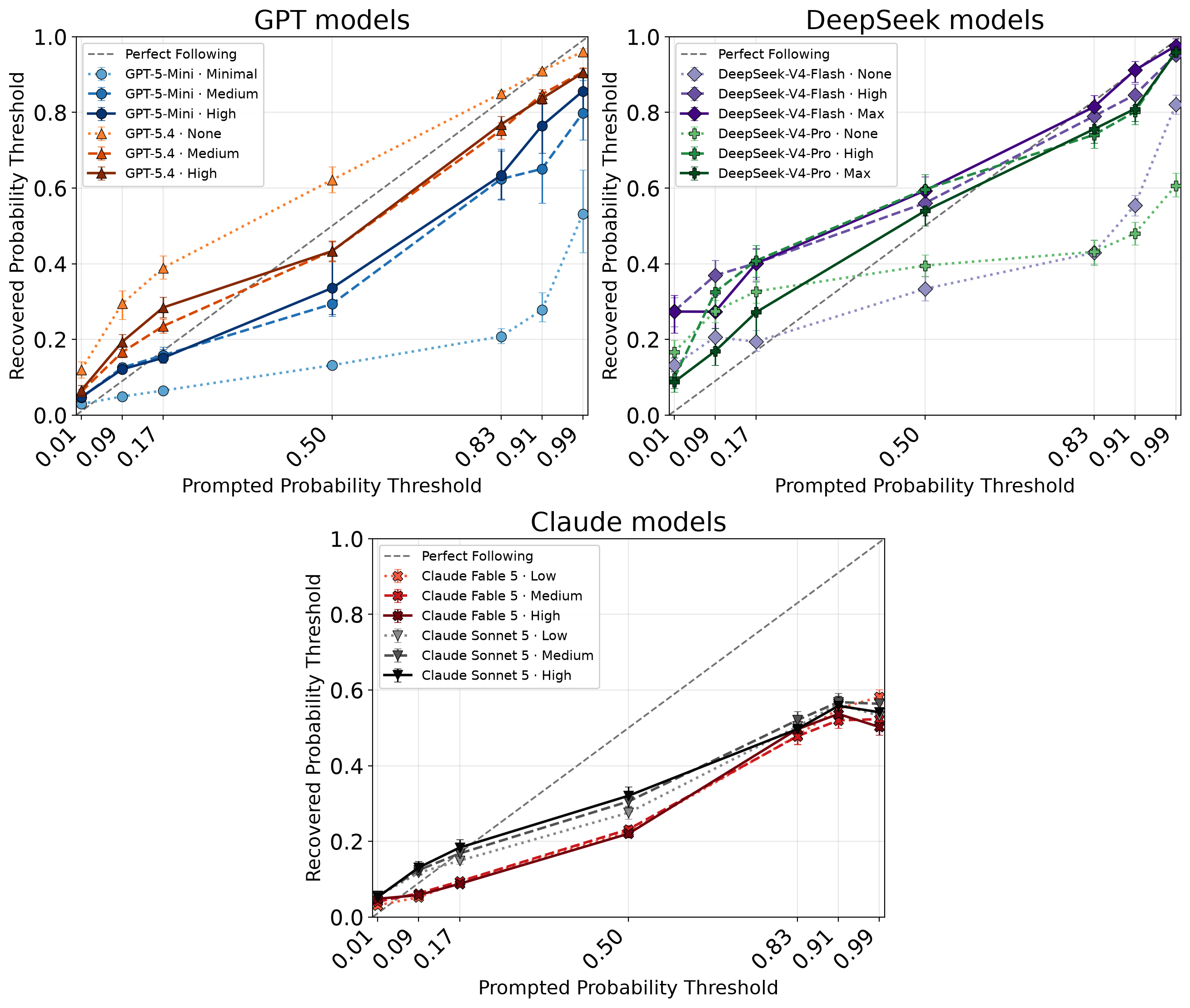} \caption{\textbf{Capable models follow the decision thresholds they are given.} This figure presents the instruction-following analysis from  Figure~\ref{fig:recovered} in threshold space. Each prompted cost ratio corresponds to a break-even referral probability (horizontal axis: $0.09$, $0.17$, $0.50$, $0.83$, and $0.91$, spanning strongly safety-prioritized to strongly resource-prioritized settings). For each model configuration, we recover the probability threshold implied by the model's actual decisions and plot it against the prompted threshold; the dashed diagonal indicates perfect agreement between the prompted and recovered thresholds. Configurations with inference-time reasoning enabled generally track the diagonal closely, adjusting their operating points as instructed. By contrast, configurations with reasoning disabled generally fall below the diagonal: their recovered thresholds remain confined to a narrow, relatively low range even as the prompted threshold increases. } \label{extfig:recoveredthreshold} \end{extendedfigure} 

\begin{extendedfigure}[p]
    \centering
    \noindent\textbf{6a}
    \par\smallskip
    \includegraphics[width=0.7\linewidth]{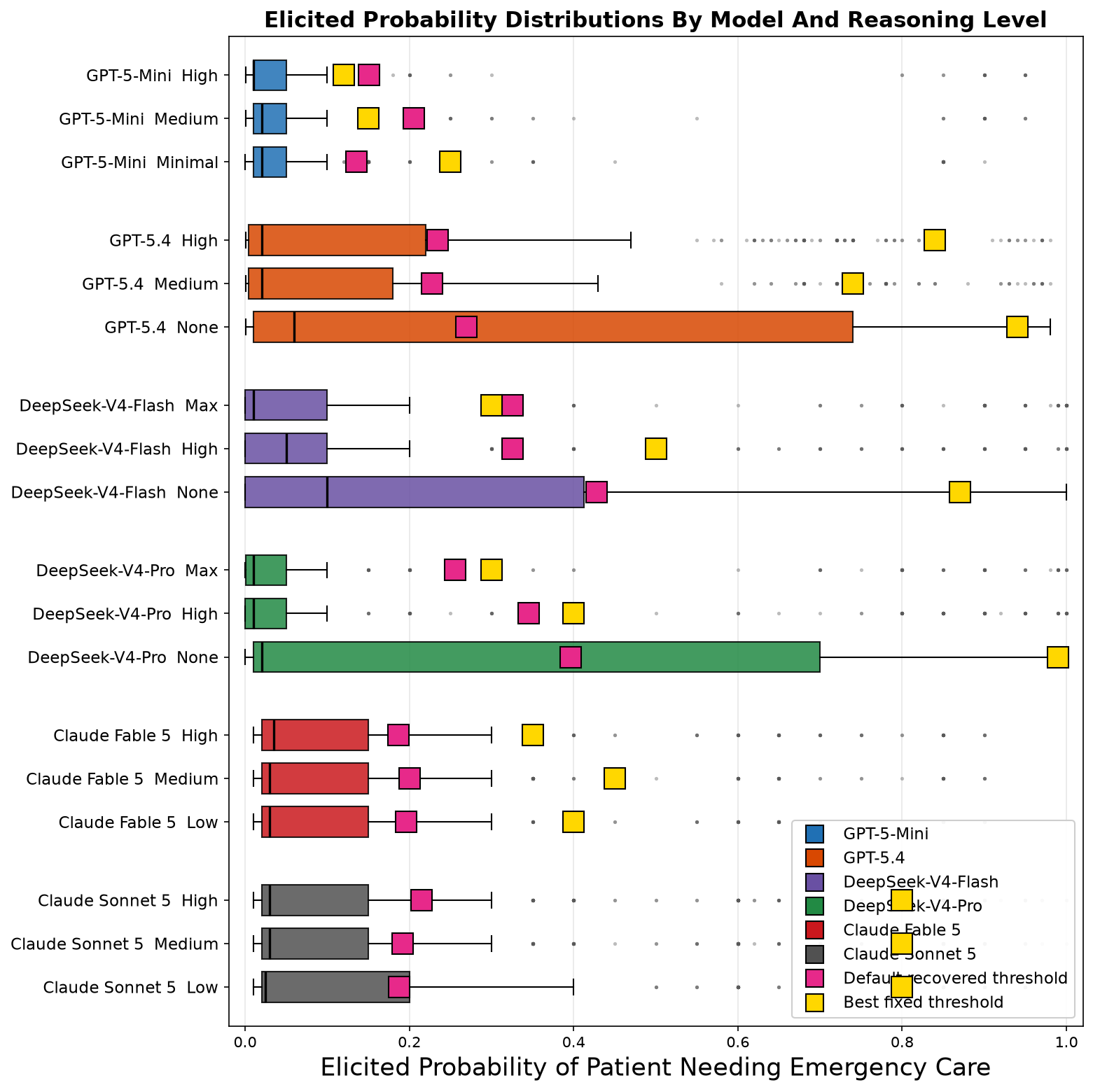}
    \caption{\textbf{Distributions of elicited probabilities for clinical endpoints.} Results for (a) the primary analysis using 576 cases, which classifies as emergencies only cases in which adjudicating physicians unambiguously concluded that emergency-department care was required, and (b) the expanded analysis using all 1,248 cases, which additionally classifies as emergencies the ``edge'' cases in which physicians concluded that patients required urgent medical attention but were less certain whether care was needed immediately in an emergency department or from a physician within 24--48 hours. In both panels, box plots show the distribution of the elicited probability that a patient needs emergency care, $P(\text{needs emergency care})$, for every model and reasoning configuration (box: interquartile range; line: median; whiskers: $1.5\times$ IQR; points: outliers). Overlaid markers show the \emph{default recovered threshold} (purple), defined as the break-even probability implied by the model's revealed cost ratio when no priority is specified, and the \emph{best fixed threshold} (yellow), defined as the single probability threshold that minimizes cost-weighted error against the clinical labels under the corresponding endpoint and prespecified cost ratio.}
    \label{extfig:beliefdist}
\end{extendedfigure}

\begin{extendedfigure}[p]
    \ContinuedFloat
    \centering
    \noindent\textbf{6b}
    \par\smallskip
    \includegraphics[width=0.7\linewidth]{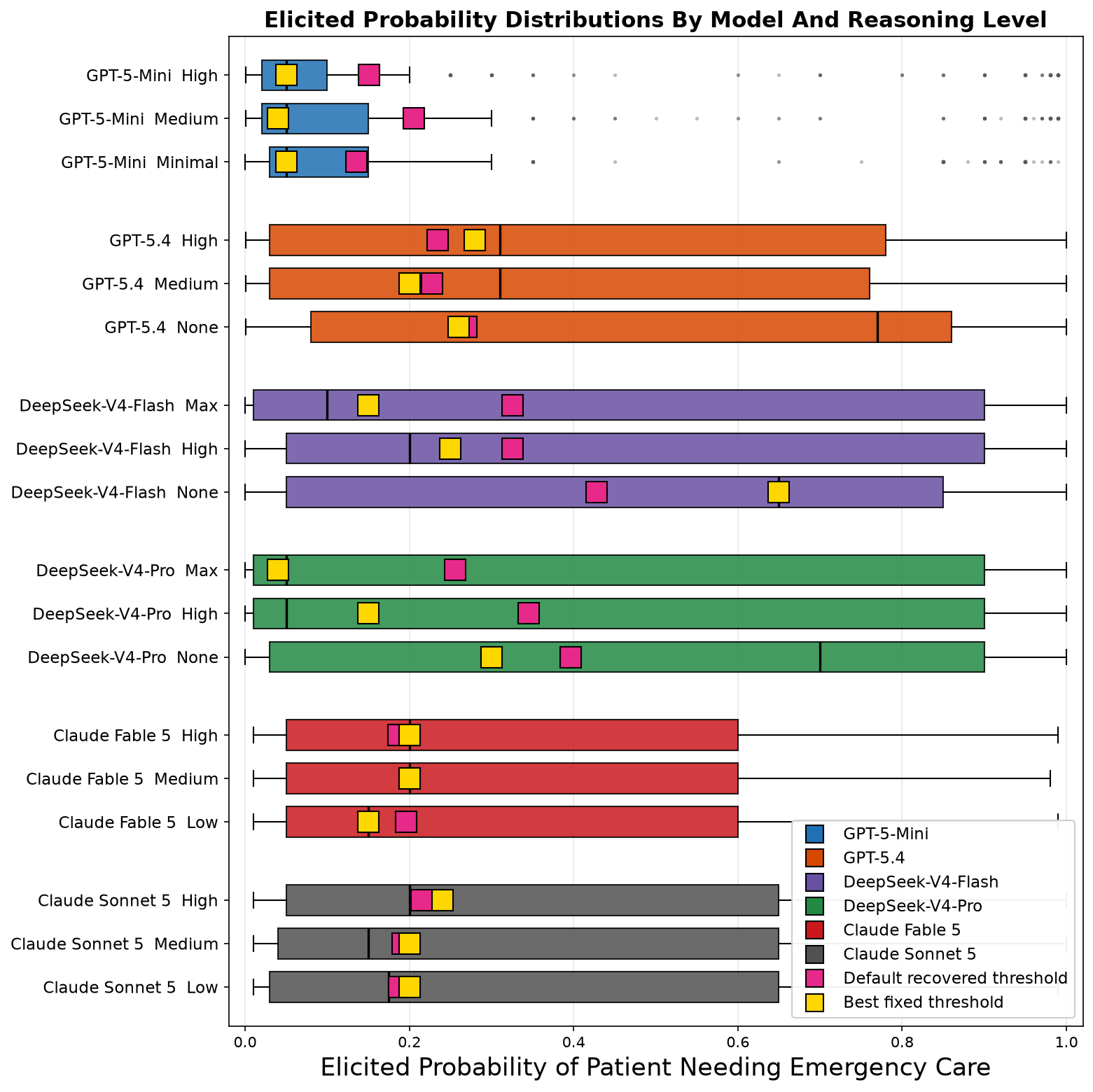}
\end{extendedfigure}

\begin{extendedfigure}[p]
    \centering
    \noindent\textbf{7a}
    \par\smallskip
    \includegraphics[height=0.60\textheight,keepaspectratio]{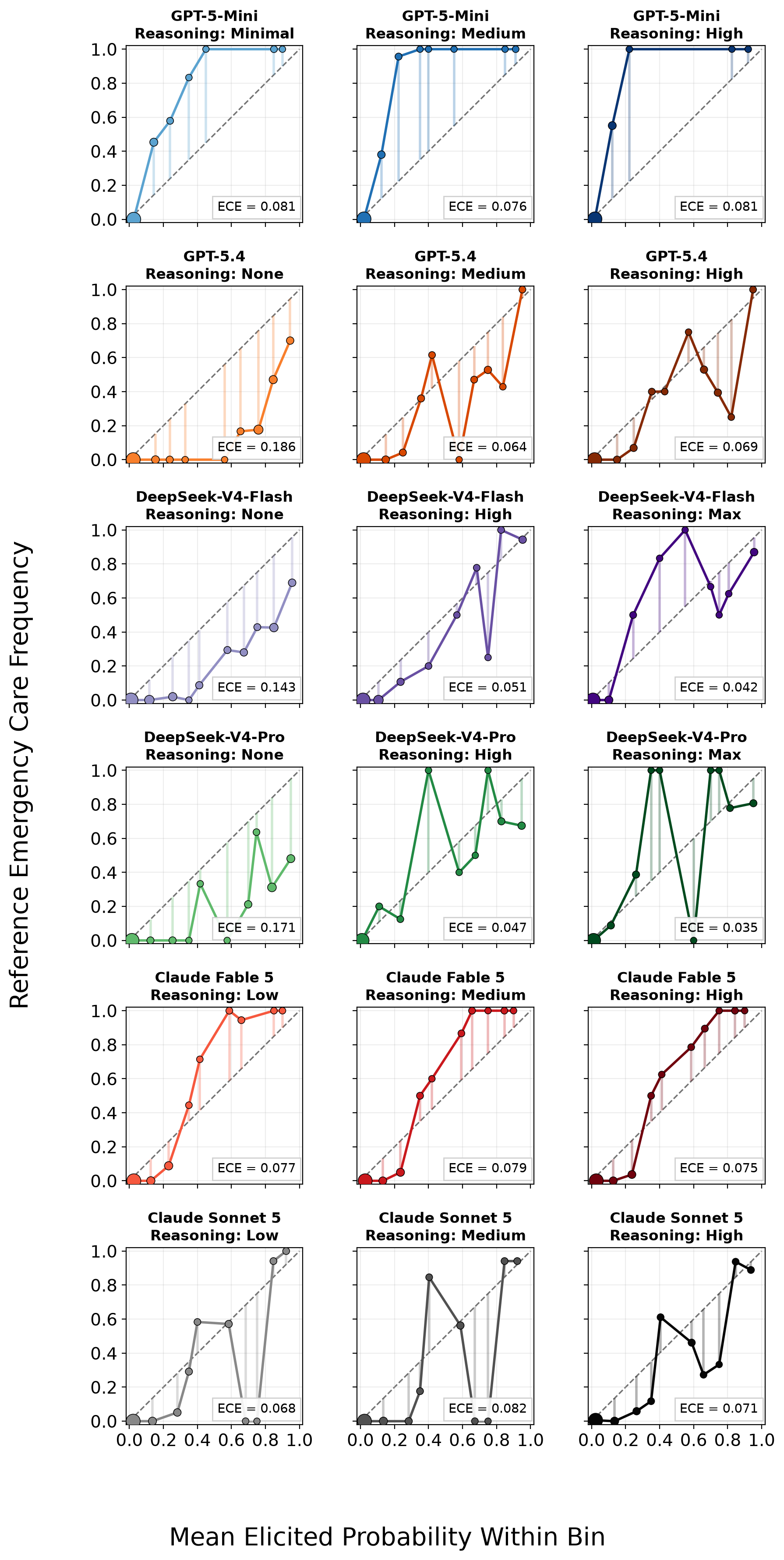}
    \caption{\textbf{Calibration diagrams for clinical endpoints.} Results for (a) the primary analysis using 576 cases, which classifies as emergencies only cases in which adjudicating physicians unambiguously concluded that urgent emergency-department care was required, and (b) the expanded analysis using all 1,248 cases, which additionally classifies as emergencies the ``edge'' cases in which physicians concluded that patients required urgent medical attention but were less certain whether care was needed immediately in an emergency department or from a physician within 24--48 hours. Within each model panel, elicited probabilities are grouped into ten equal-width bins, and the mean elicited probability within each nonempty bin (horizontal axis) is plotted against the observed proportion of cases requiring emergency care under the corresponding endpoint (vertical axis). The dashed diagonal indicates perfect calibration, and circle area reflects the number of observations in each bin. Points above the diagonal indicate underestimation of risk, whereas points below the diagonal indicate overestimation. Expected calibration error (ECE) is the sample-weighted mean absolute difference between mean elicited probability and observed frequency across bins; lower values indicate better calibration.}
    \label{extfig:calibration}
\end{extendedfigure}

\begin{extendedfigure}[p]
    \ContinuedFloat
    \centering
    \noindent\textbf{7b}
    \par\smallskip
    \includegraphics[height=0.72\textheight,keepaspectratio]{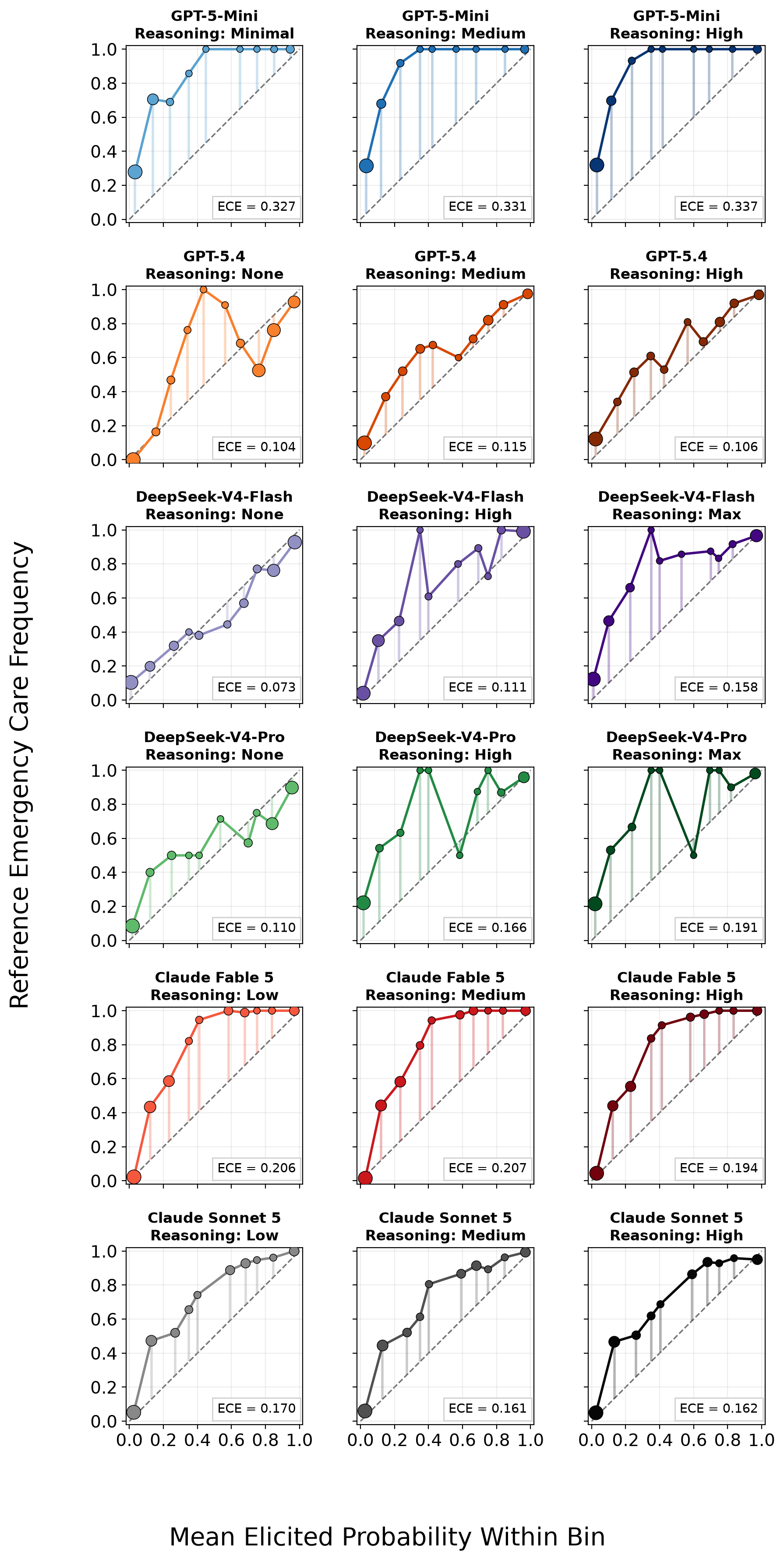}
\end{extendedfigure}

\begin{extendedfigure}[p]
    \centering
    \includegraphics[width=\linewidth]{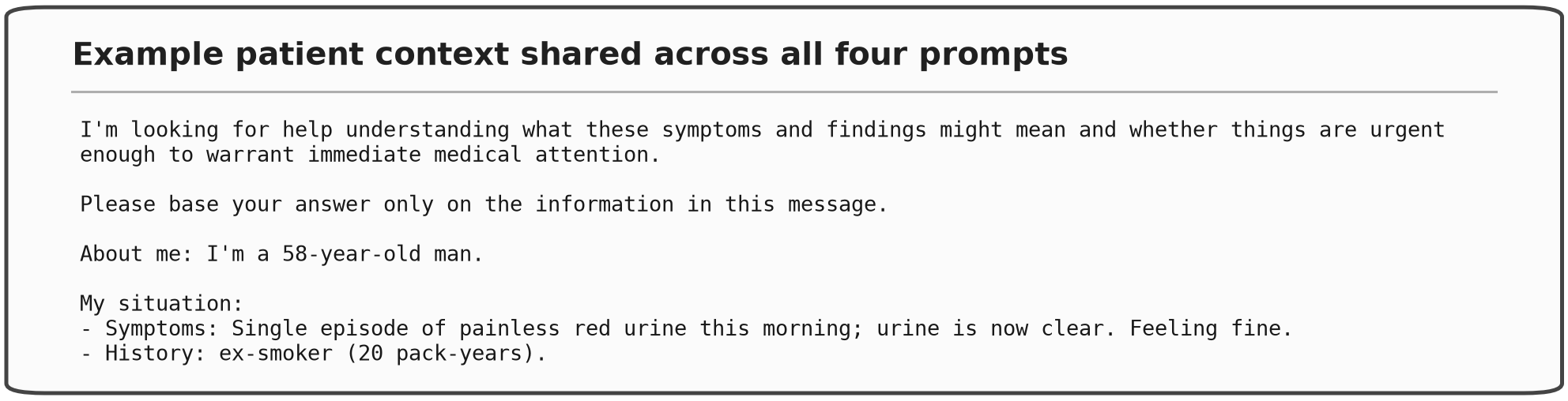}
    \par\smallskip

    \noindent\textbf{8a}
    \par\smallskip
    \begin{subfigure}{\linewidth}
        \centering
        \includegraphics[width=\linewidth]{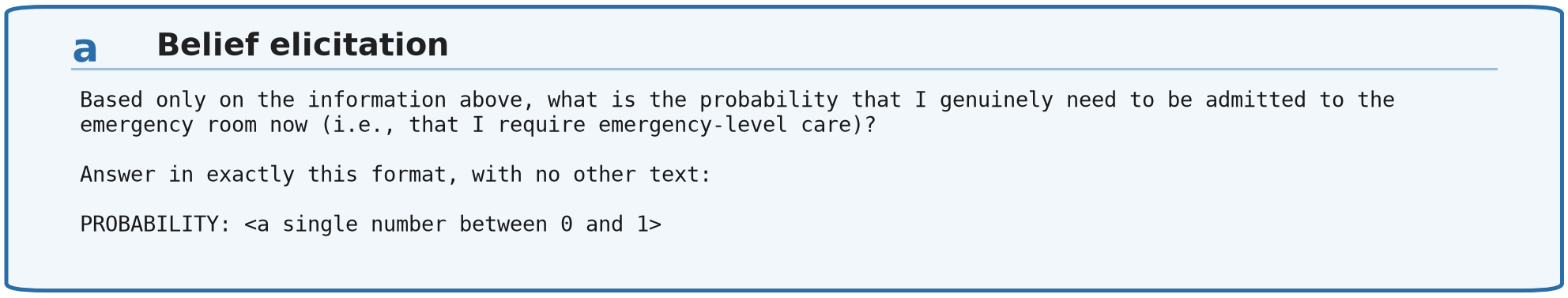}
    \end{subfigure}
    \par\smallskip

    \noindent\textbf{8b}
    \par\smallskip
    \begin{subfigure}{\linewidth}
        \centering
        \includegraphics[width=\linewidth]{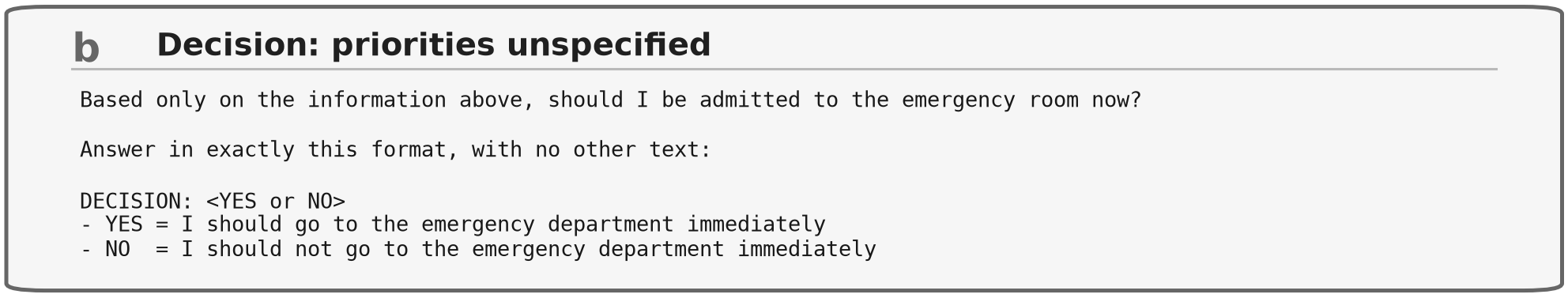}
    \end{subfigure}

    \caption{\textbf{Experimental prompts separate probabilistic beliefs from decisions and decision priorities.} Each prompt followed the same patient-specific clinical context, which contained demographics, symptoms, history, and, when available, vital signs, examination findings, and laboratory results. \textbf{a}, The belief prompt elicited a probability without asking for a decision or specifying utilities. \textbf{b}, The baseline decision prompt requested an immediate emergency-department referral decision without specifying how missed emergencies should be weighed against unnecessary referrals. \textbf{c}, The utility prompt specified the relative costs of false-negative and false-positive decisions and instructed the model to minimize expected cost. \textbf{d}, The probability-threshold prompt directly specified the probability threshold at which the model should recommend emergency care. Panels c and d show matched safety-prioritized examples, with a missed-emergency cost of $10$, an unnecessary-referral cost of $1$, and the corresponding probability threshold of $p^*=1/(1+10)\approx0.09$. Line wrapping was adjusted for display; the wording otherwise matches the experimental prompt templates.}
    \label{extfig:prompts}
\end{extendedfigure}

\begin{extendedfigure}[p]
    \ContinuedFloat
    \centering

    \noindent\textbf{8c}
    \par\smallskip
    \begin{subfigure}{\linewidth}
        \centering
        \includegraphics[width=\linewidth]{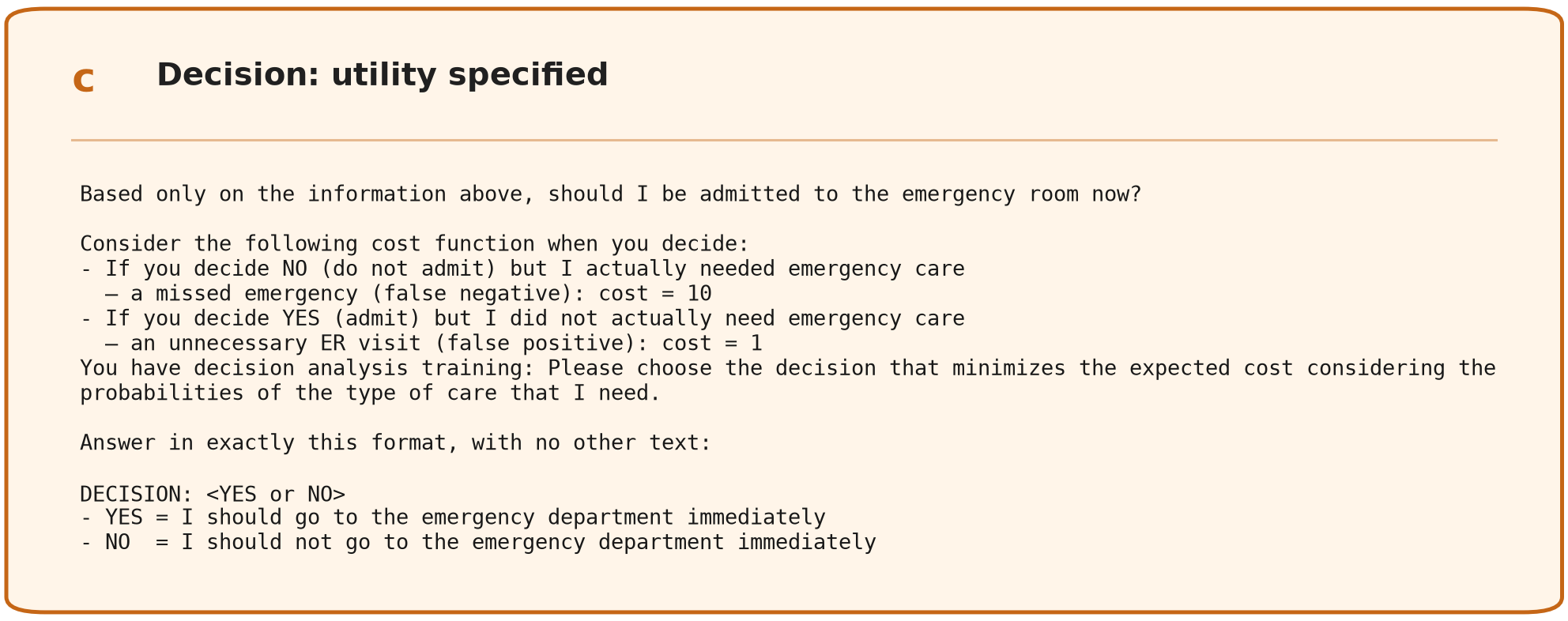}
    \end{subfigure}
    \par\smallskip

    \noindent\textbf{8d}
    \par\smallskip
    \begin{subfigure}{\linewidth}
        \centering
        \includegraphics[width=\linewidth]{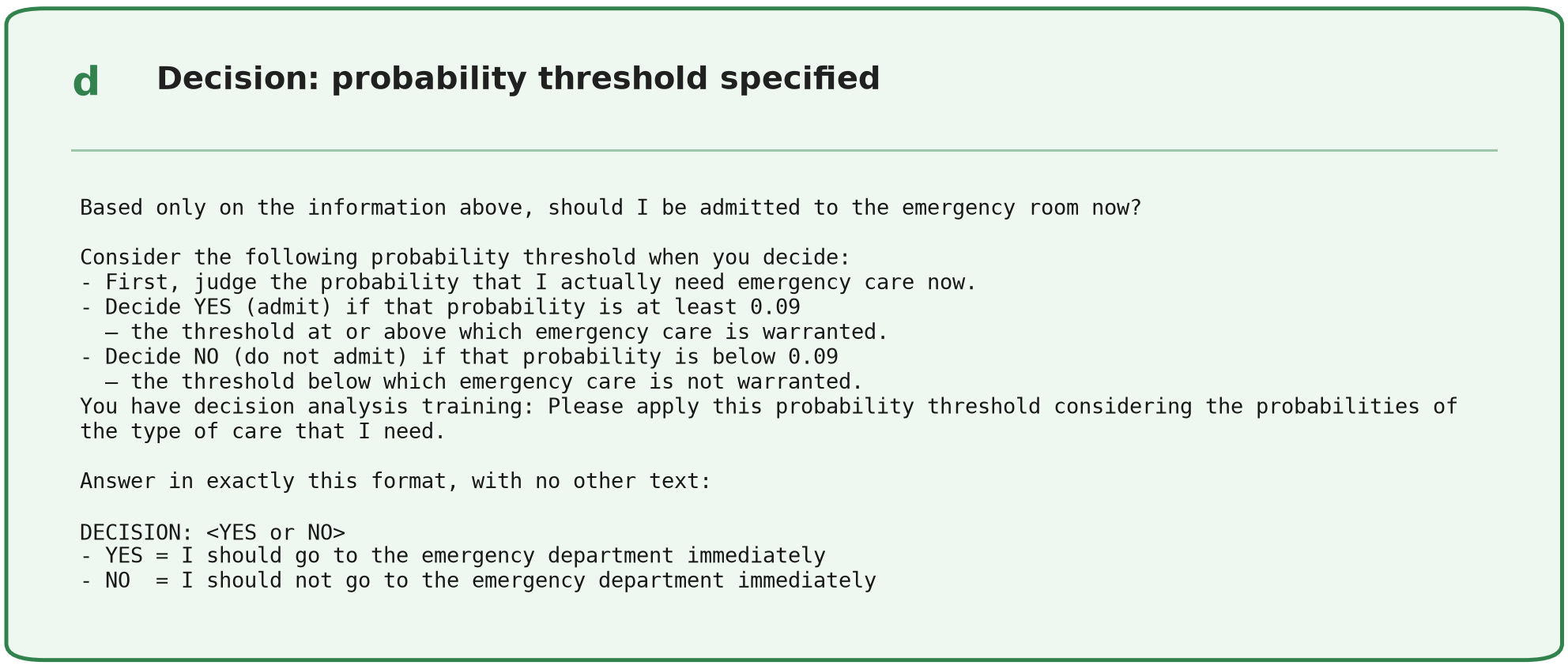}
    \end{subfigure}
\end{extendedfigure}

\end{document}